\documentclass{article}

\usepackage[preprint,nonatbib]{neurips_2023}
\usepackage[numbers]{natbib}

\usepackage[utf8]{inputenc} %
\usepackage[T1]{fontenc}    %
\usepackage{url}            %
\usepackage{booktabs}       %
\usepackage{amsfonts}       %
\usepackage{nicefrac}       %
\usepackage{microtype}      %
\usepackage[dvipsnames]{xcolor}         %

\usepackage[pagebackref=true,breaklinks=true,colorlinks,bookmarks=true]{hyperref}
\usepackage[titletoc,title]{appendix}

\title{Fast High-Resolution Image Synthesis with\\Latent Adversarial Diffusion Distillation}

\author{
\textbf{Axel Sauer} \quad \textbf{Frederic Boesel} \quad \textbf{Tim Dockhorn} \\\\
\textbf{Andreas Blattmann} \quad \textbf{Patrick Esser} \quad \textbf{Robin Rombach} \\\\
Stability AI
\vspace{-2em}
}

\newcommand{\editcomparison}{
\begin{figure*}[t]
\centering
\scriptsize
\resizebox{\linewidth}{!}{%
\begin{tabular}{@{\hspace{0\tabcolsep}}c@{\hspace{0.2\tabcolsep}}c@{\hspace{0.2\tabcolsep}}c@{\hspace{0.2\tabcolsep}}c@{\hspace{0.2\tabcolsep}}c@{\hspace{0.2\tabcolsep}}c@{\hspace{0.2\tabcolsep}}c}
& Input & InstructPix2Pix~\citep{brooks2023instructpix2pix} & Magicbrush~\citep{zhang2024magicbrush} & Hive~\citep{zhang2023hive} & SD3-edit &  SD3-edit Turbo \\
{
\begin{tabular}[x]{@{}c@{}}  Change the dog\\to a wolf \end{tabular}}
 & 
\raisebox{-.5\height}{
\includegraphics[width=0.15\linewidth]{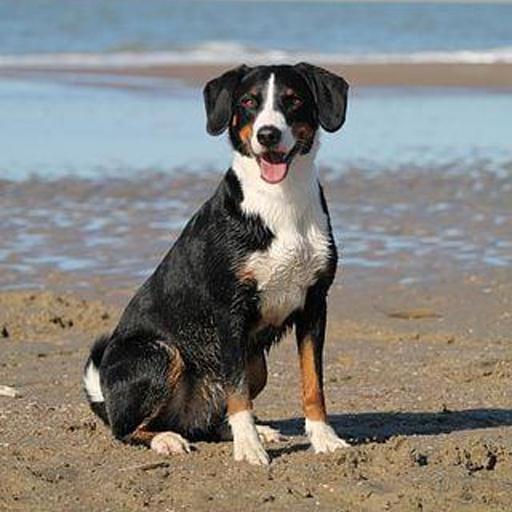}}&
\raisebox{-.5\height}{
\includegraphics[width=0.15\linewidth]{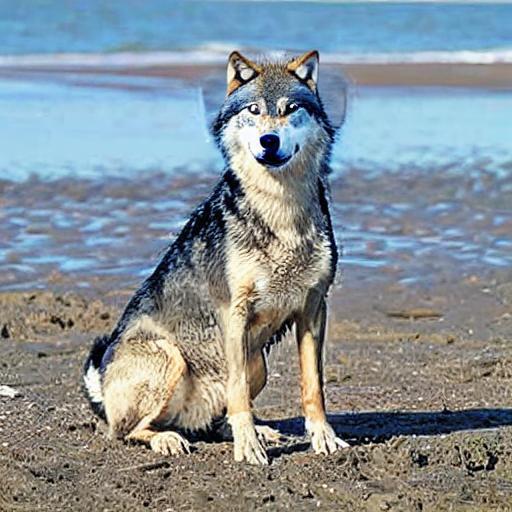}}&
\raisebox{-.5\height}{
\includegraphics[width=0.15\linewidth]{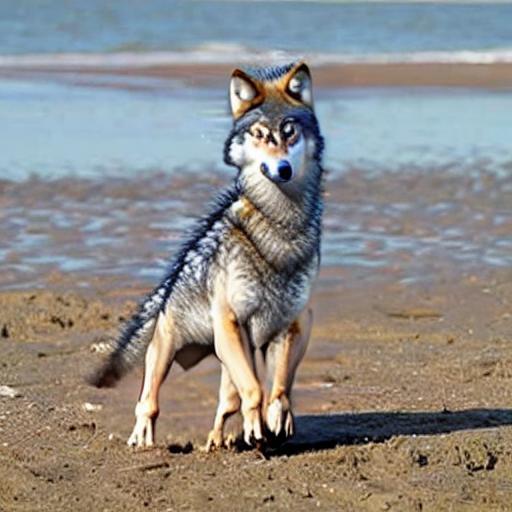}}&
\raisebox{-.5\height}{
\includegraphics[width=0.15\linewidth]{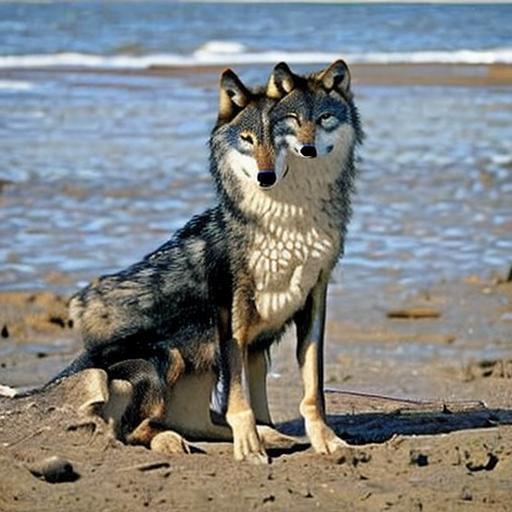}}&
\raisebox{-.5\height}{
\includegraphics[width=0.15\linewidth]{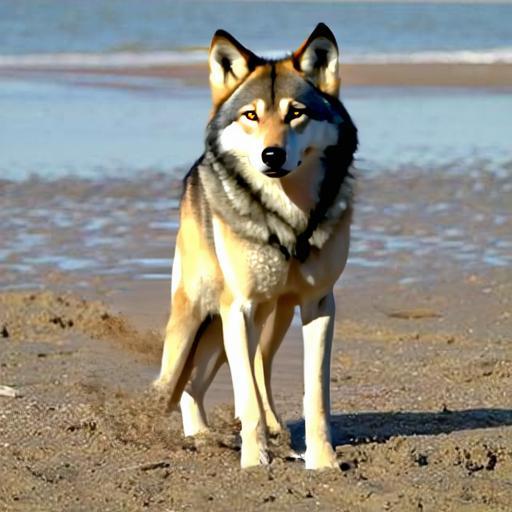}} &
\raisebox{-.5\height}{
\includegraphics[width=0.15\linewidth]{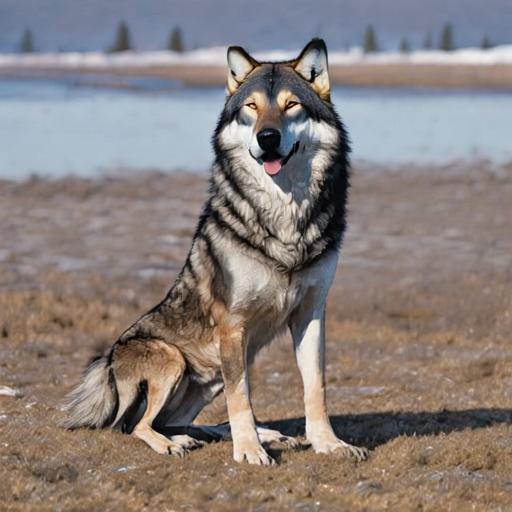}}\\

{
\begin{tabular}[x]{@{}c@{}}  Add earrings to\\the woman \end{tabular}}
 & 
\raisebox{-.5\height}{
\includegraphics[width=0.15\linewidth]{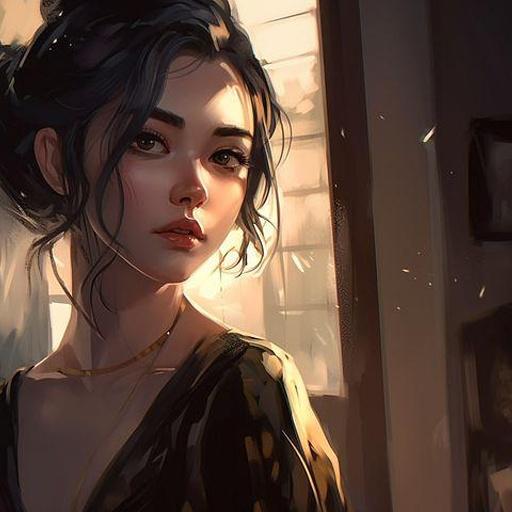}}&
\raisebox{-.5\height}{
\includegraphics[width=0.15\linewidth]{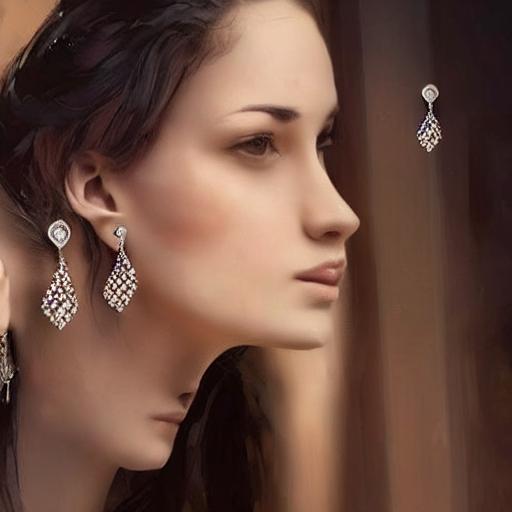}}&
\raisebox{-.5\height}{
\includegraphics[width=0.15\linewidth]{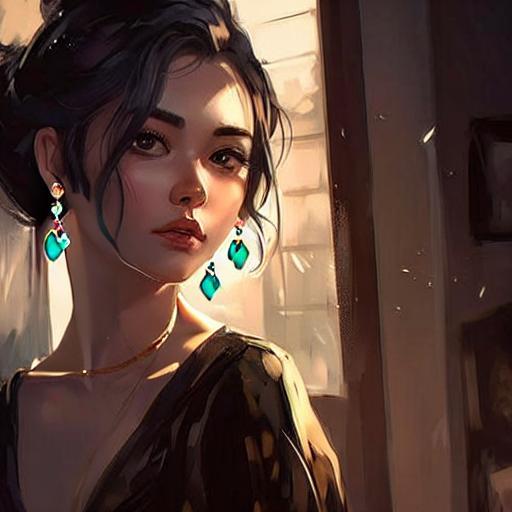}}&
\raisebox{-.5\height}{
\includegraphics[width=0.15\linewidth]{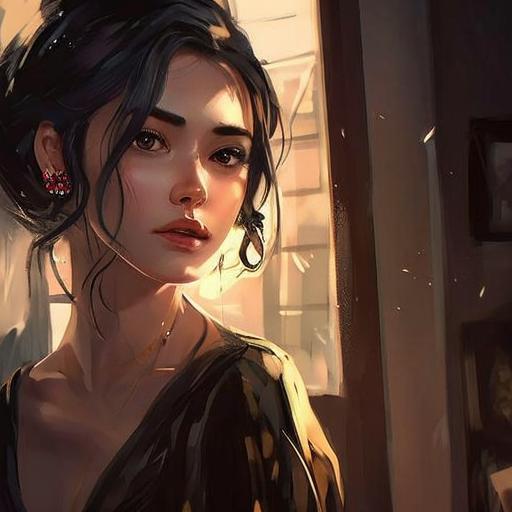}}&
\raisebox{-.5\height}{
\includegraphics[width=0.15\linewidth]{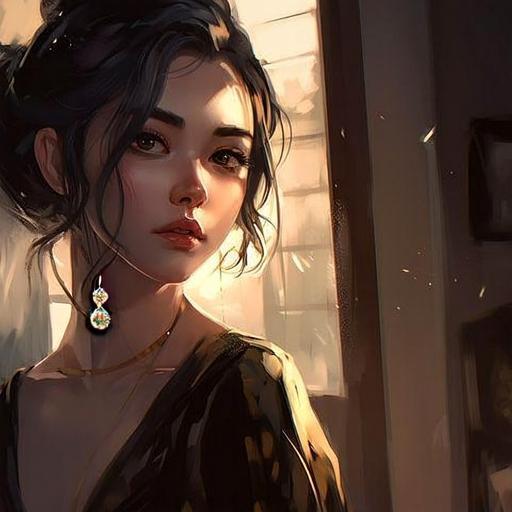}} &
\raisebox{-.5\height}{
\includegraphics[width=0.15\linewidth]{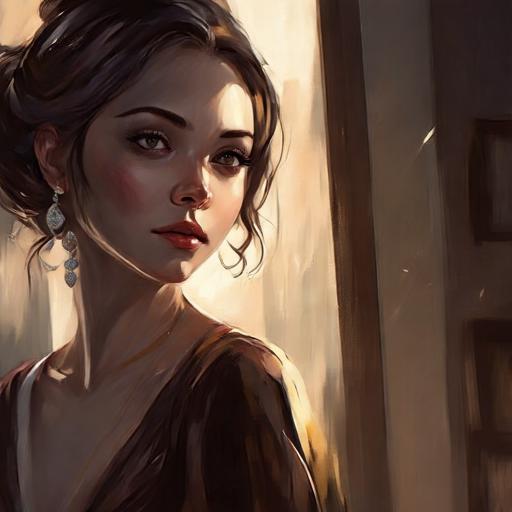}}\\

{
\begin{tabular}[x]{@{}c@{}}  Change to\\pixel art style \end{tabular}}
 & 
\raisebox{-.5\height}{
\includegraphics[width=0.15\linewidth]{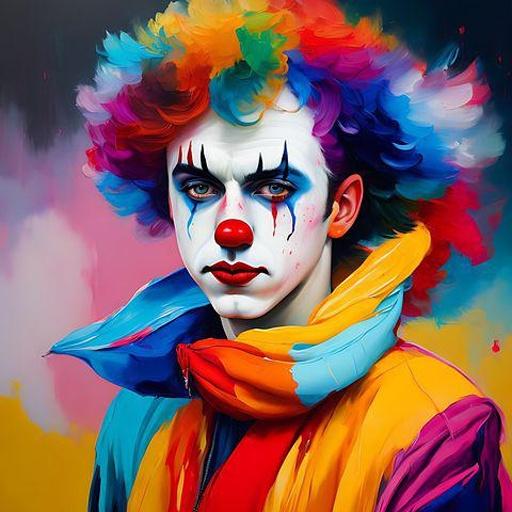}}&
\raisebox{-.5\height}{
\includegraphics[width=0.15\linewidth]{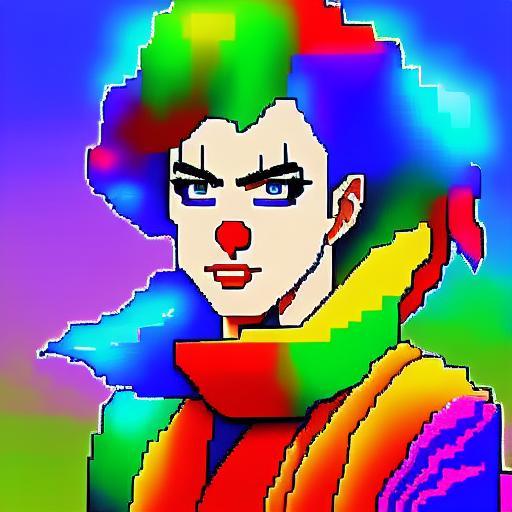}}&
\raisebox{-.5\height}{
\includegraphics[width=0.15\linewidth]{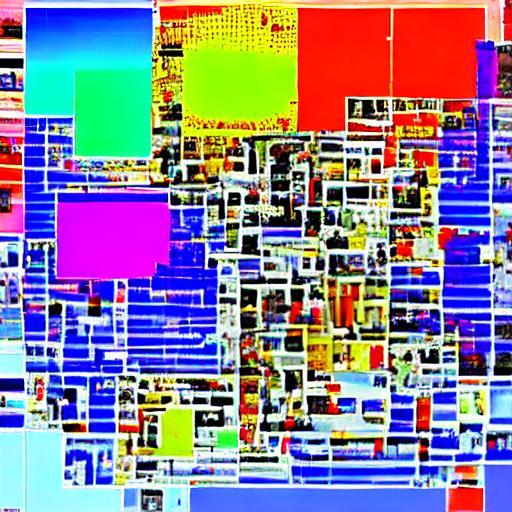}}&
\raisebox{-.5\height}{
\includegraphics[width=0.15\linewidth]{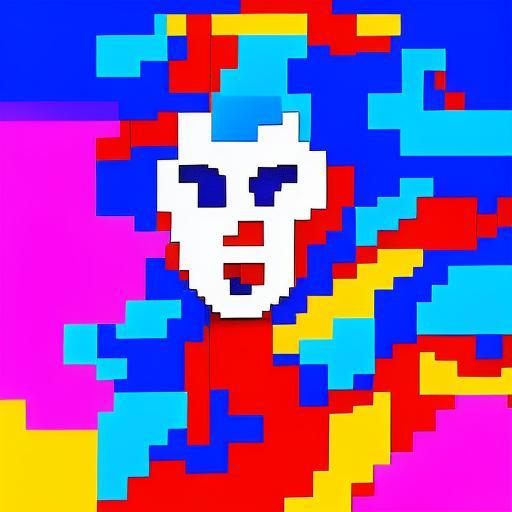}}&
\raisebox{-.5\height}{
\includegraphics[width=0.15\linewidth]{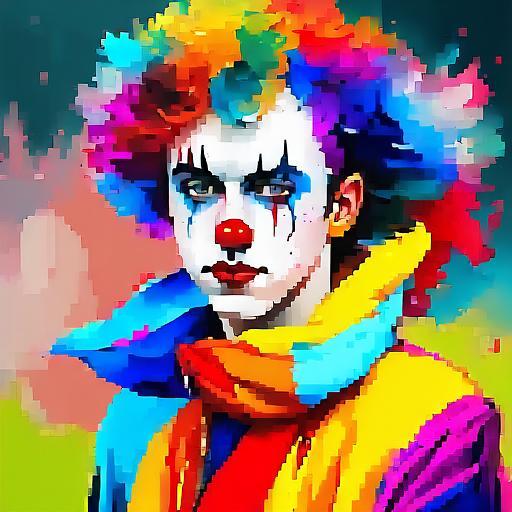}} &
\raisebox{-.5\height}{
\includegraphics[width=0.15\linewidth]{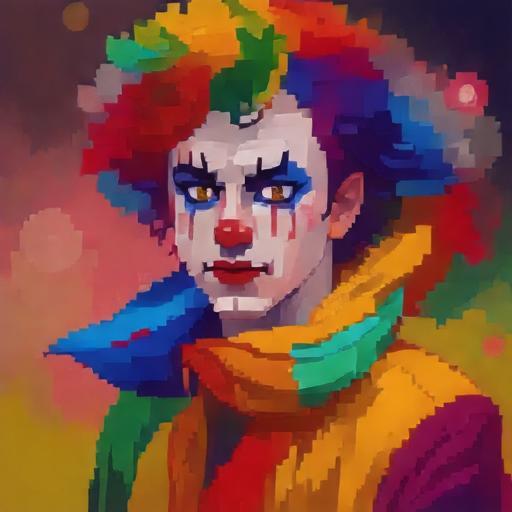}}\\

{
\begin{tabular}[x]{@{}c@{}}  Change the\\ animal from a cat \\into a tiger \end{tabular}}
 & 
\raisebox{-.5\height}{
\includegraphics[width=0.15\linewidth]{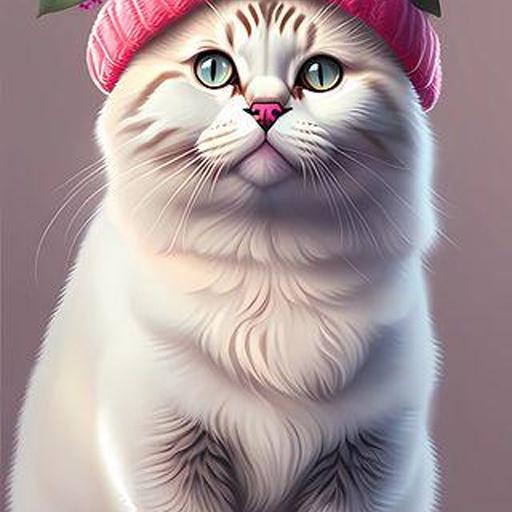}}&
\raisebox{-.5\height}{
\includegraphics[width=0.15\linewidth]{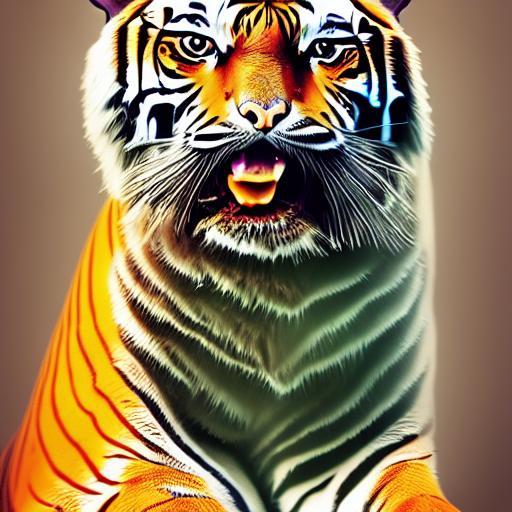}}&
\raisebox{-.5\height}{
\includegraphics[width=0.15\linewidth]{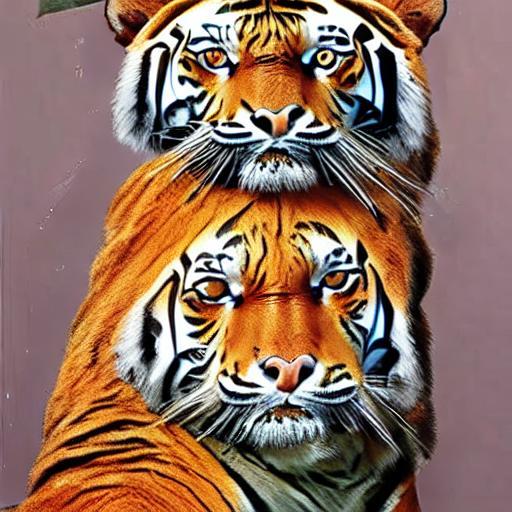}}&
\raisebox{-.5\height}{
\includegraphics[width=0.15\linewidth]{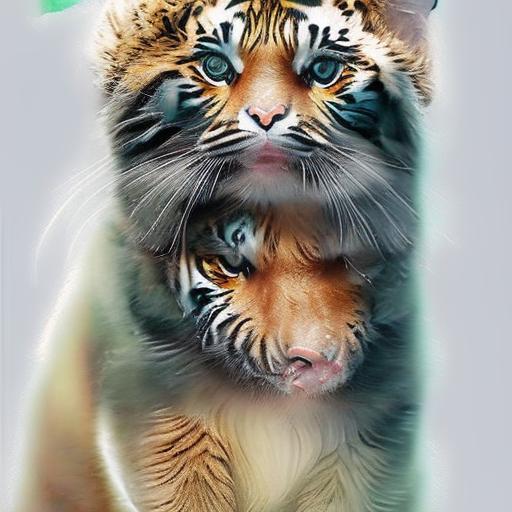}}&
\raisebox{-.5\height}{
\includegraphics[width=0.15\linewidth]{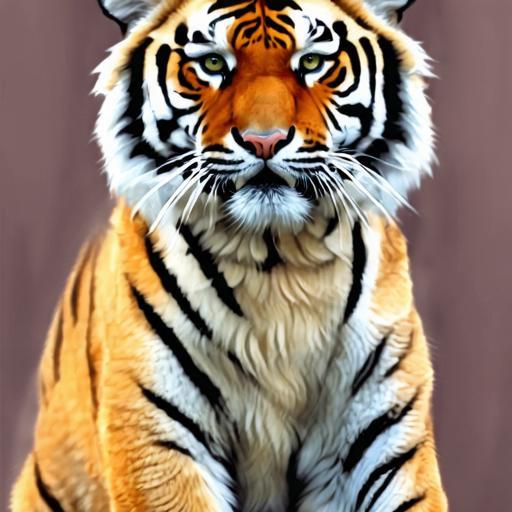}} &
\raisebox{-.5\height}{
\includegraphics[width=0.15\linewidth]{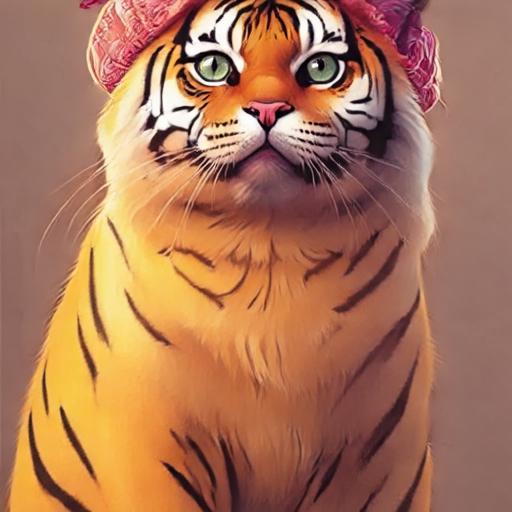}}\\

{
\begin{tabular}[x]{@{}c@{}}  Replace the\\ dog with \\a monkey \end{tabular}}
 & 
\raisebox{-.5\height}{
\includegraphics[width=0.15\linewidth]{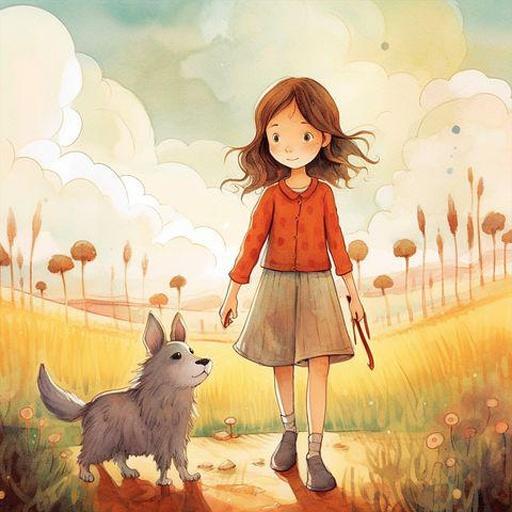}}&
\raisebox{-.5\height}{
\includegraphics[width=0.15\linewidth]{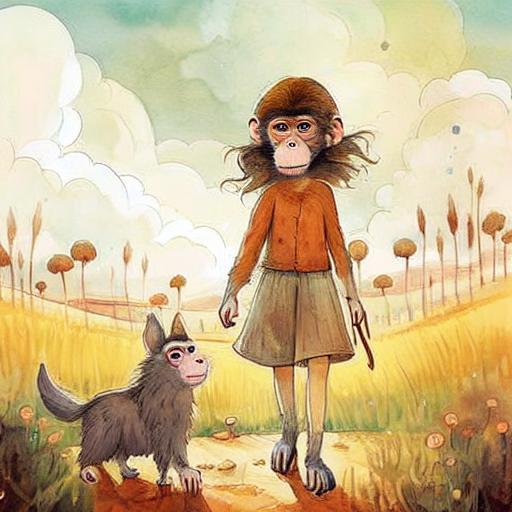}}&
\raisebox{-.5\height}{
\includegraphics[width=0.15\linewidth]{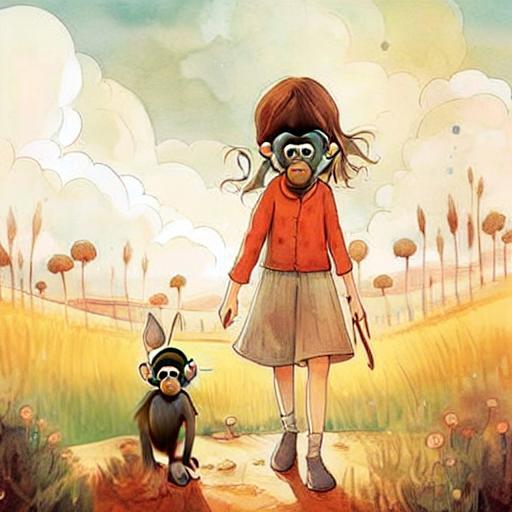}}&
\raisebox{-.5\height}{
\includegraphics[width=0.15\linewidth]{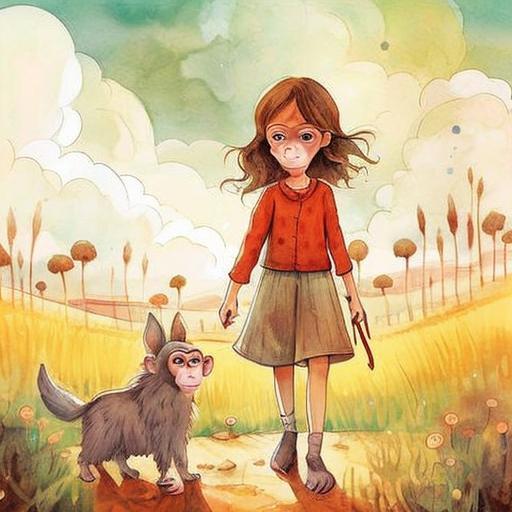}}&
\raisebox{-.5\height}{
\includegraphics[width=0.15\linewidth]{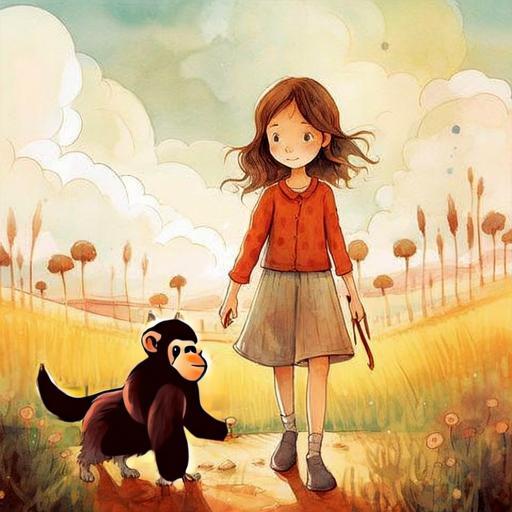}} &
\raisebox{-.5\height}{
\includegraphics[width=0.15\linewidth]{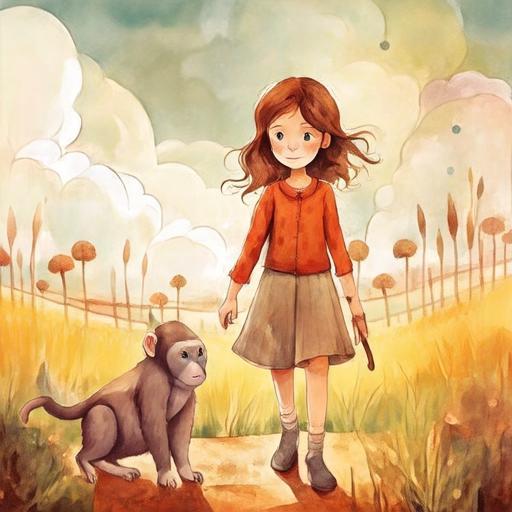}}\\

{
\begin{tabular}[x]{@{}c@{}}  Change the plant\\ to a flower \end{tabular}}
 & 
\raisebox{-.5\height}{
\includegraphics[width=0.15\linewidth]{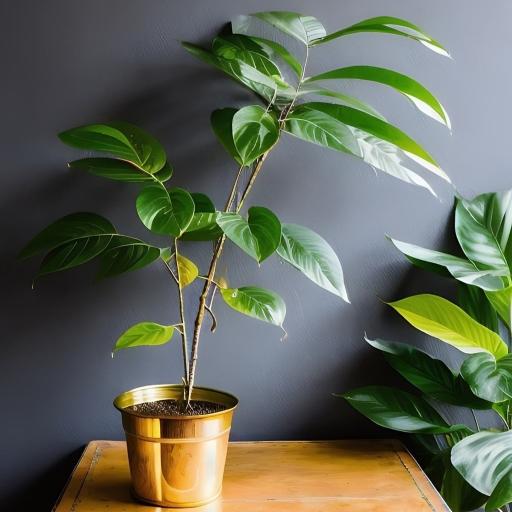}}&
\raisebox{-.5\height}{
\includegraphics[width=0.15\linewidth]{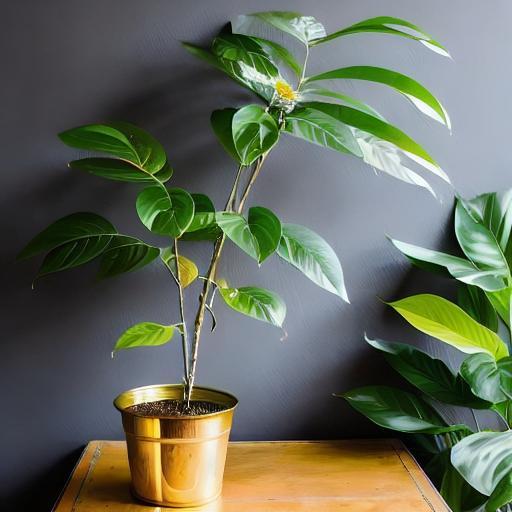}}&
\raisebox{-.5\height}{
\includegraphics[width=0.15\linewidth]{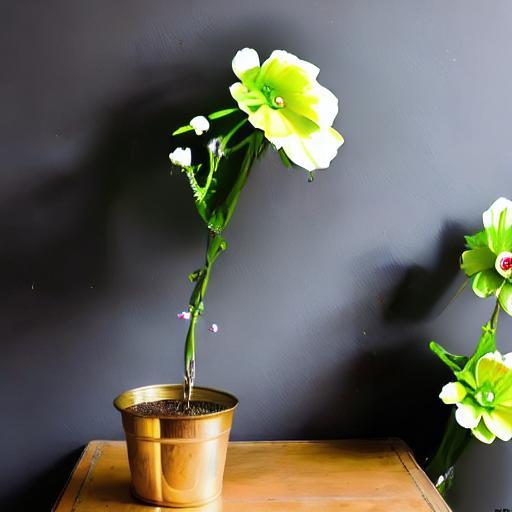}}&
\raisebox{-.5\height}{
\includegraphics[width=0.15\linewidth]{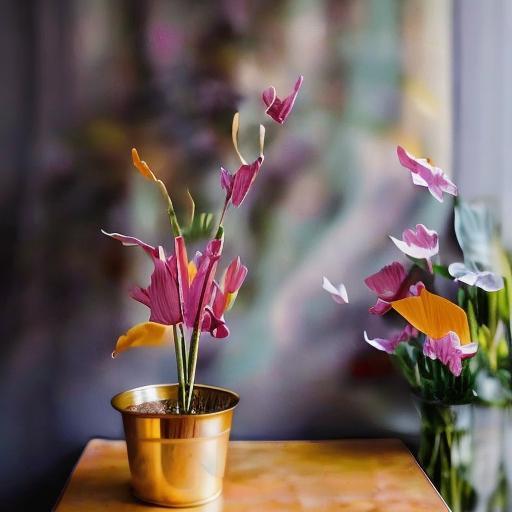}}&
\raisebox{-.5\height}{
\includegraphics[width=0.15\linewidth]{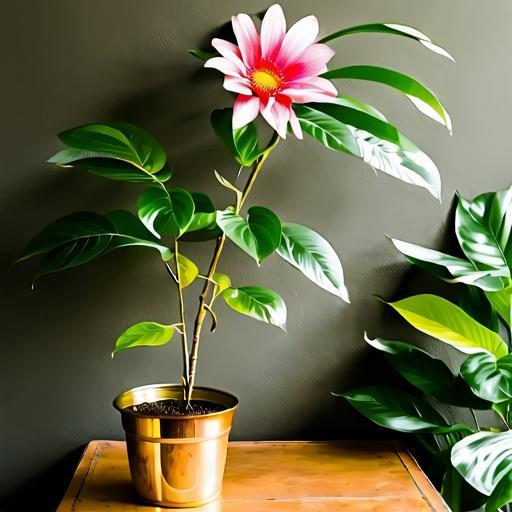}} &
\raisebox{-.5\height}{
\includegraphics[width=0.15\linewidth]{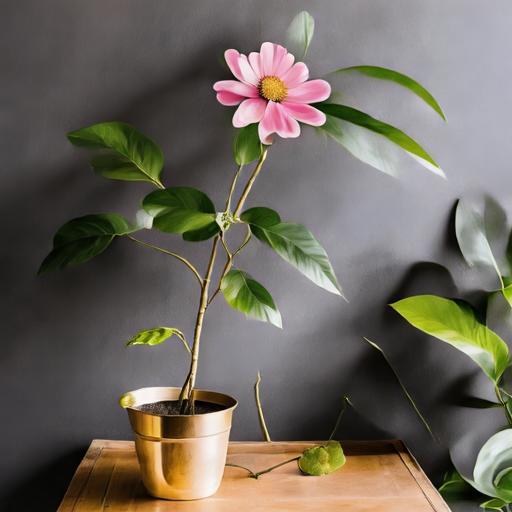}}\\
\end{tabular}
}
\caption{
\textbf{Qualitative comparison for instruction-based editing.} 
For a given prompt and input image, we compare our distilled SD3-edit Turbo (1 step) to its teacher SD3-edit (50 steps) and several other baselines.
}
\label{fig:edit_comparison} %
\end{figure*}
}

\newcommand{\inpaintcomparison}{
\begin{figure*}[t]
\centering
\scriptsize
\begin{tabular}{@{\hspace{0\tabcolsep}}c@{\hspace{0.\tabcolsep}}c@{\hspace{0.\tabcolsep}}c@{\hspace{0.\tabcolsep}}c@{\hspace{0.\tabcolsep}}c@{\hspace{0.\tabcolsep}}c}
Original Input & Masked Input & LaMa &  SD1.5 inpainting & SD3-inpainting  & SD3-inpainting Turbo\vspace{0.3em}\\
\raisebox{-.5\height}{
\includegraphics[width=0.15\linewidth]{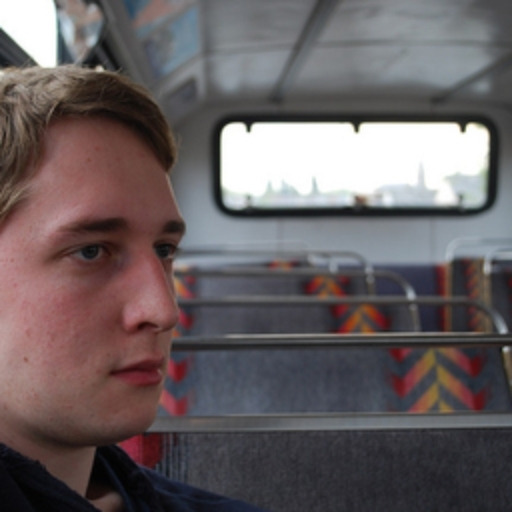}}&
\raisebox{-.5\height}{
\includegraphics[width=0.15\linewidth]{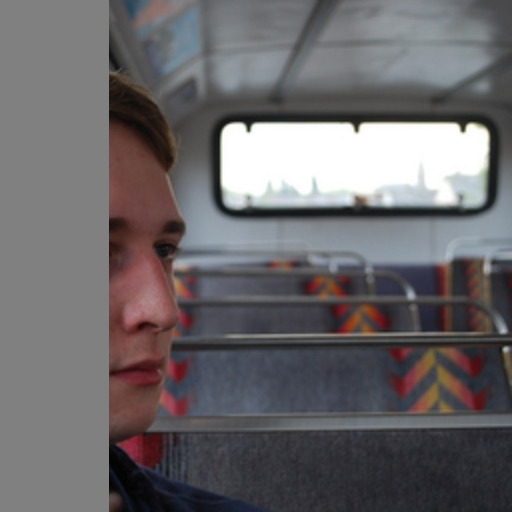}}&
\raisebox{-.5\height}{
\includegraphics[width=0.15\linewidth]{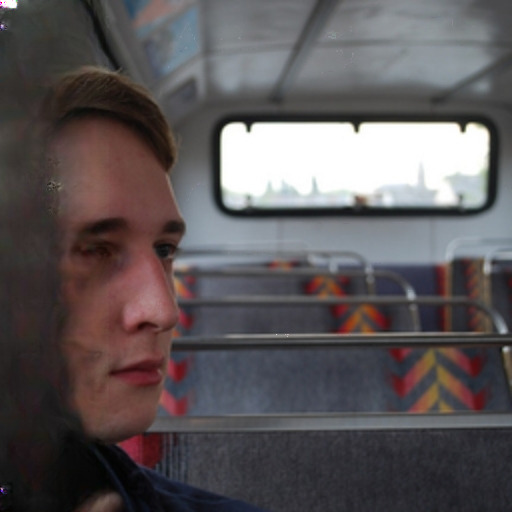}}&
\raisebox{-.5\height}{
\includegraphics[width=0.15\linewidth]{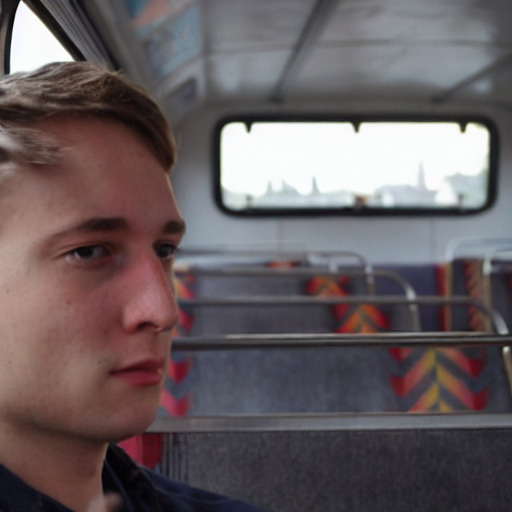}}&
\raisebox{-.5\height}{
\includegraphics[width=0.15\linewidth]{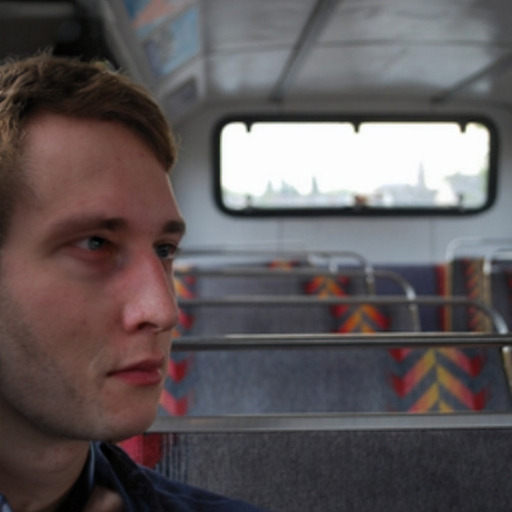}}&
\raisebox{-.5\height}{
\includegraphics[width=0.15\linewidth]{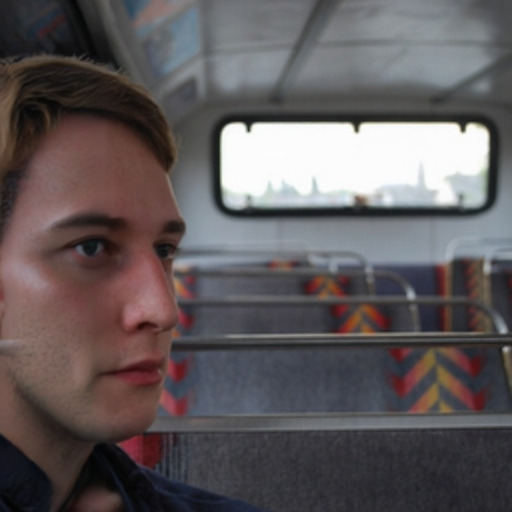}} \\

\raisebox{-.5\height}{
\includegraphics[width=0.15\linewidth]{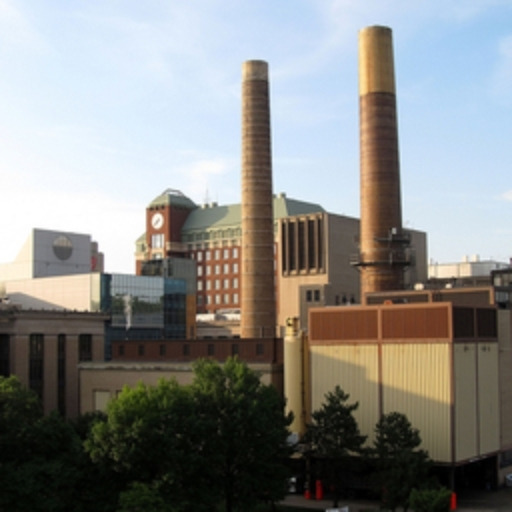}}&
\raisebox{-.5\height}{
\includegraphics[width=0.15\linewidth]{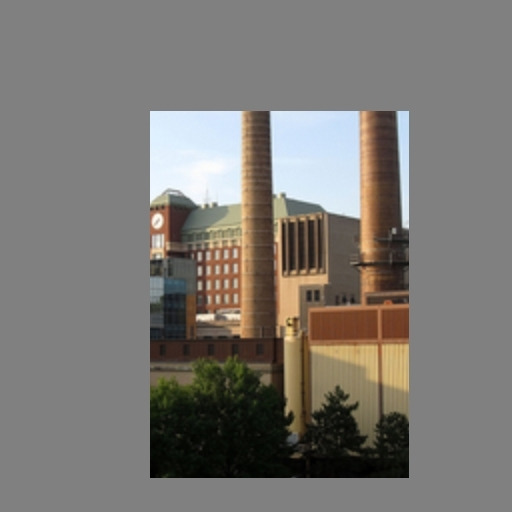}}&
\raisebox{-.5\height}{
\includegraphics[width=0.15\linewidth]{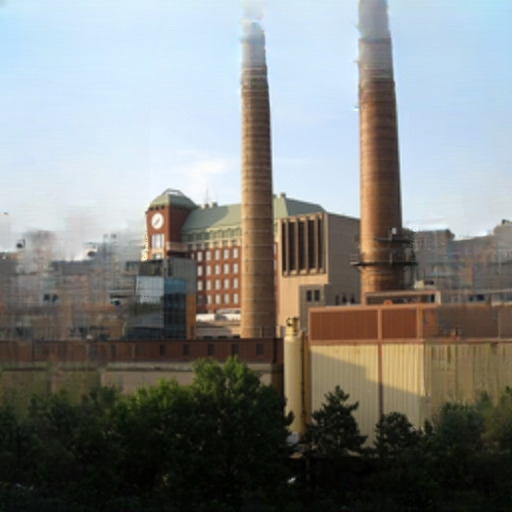}}&
\raisebox{-.5\height}{
\includegraphics[width=0.15\linewidth]{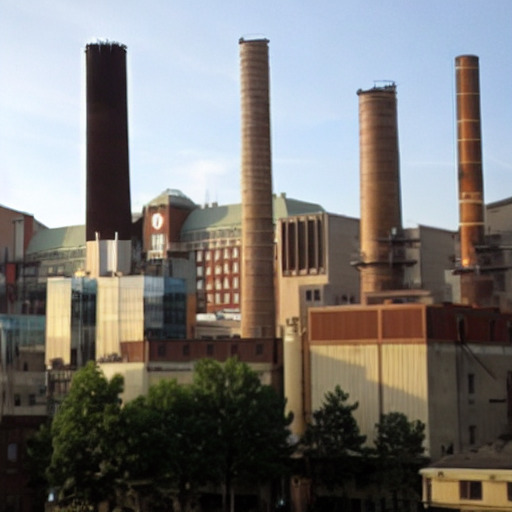}}&
\raisebox{-.5\height}{
\includegraphics[width=0.15\linewidth]{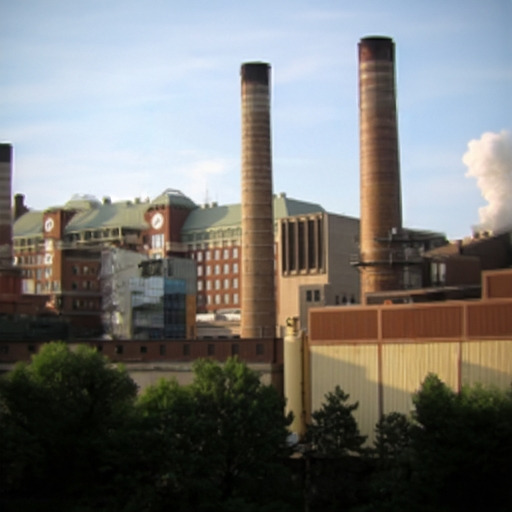}}&
\raisebox{-.5\height}{
\includegraphics[width=0.15\linewidth]{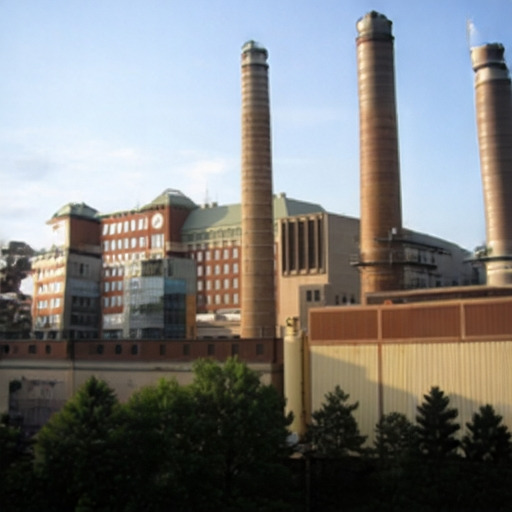}} \\

\raisebox{-.5\height}{
\includegraphics[width=0.15\linewidth]{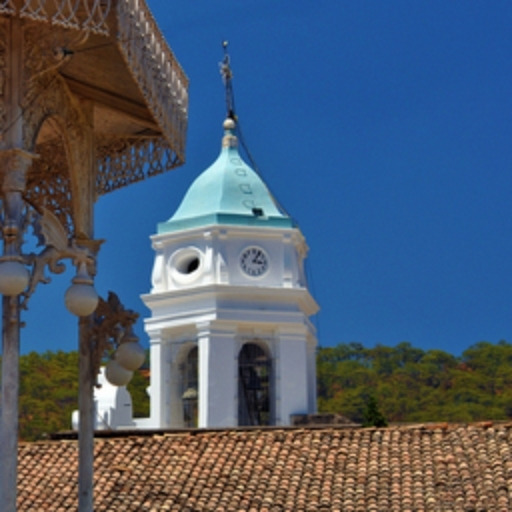}}&
\raisebox{-.5\height}{
\includegraphics[width=0.15\linewidth]{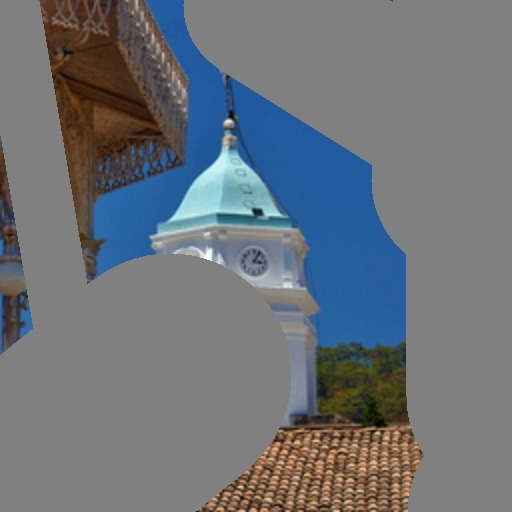}}&
\raisebox{-.5\height}{
\includegraphics[width=0.15\linewidth]{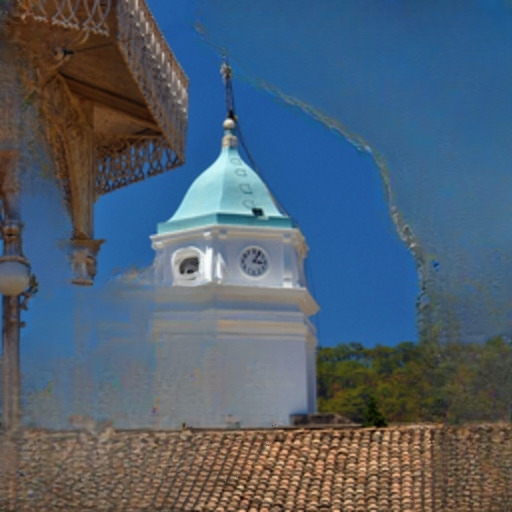}}&
\raisebox{-.5\height}{
\includegraphics[width=0.15\linewidth]{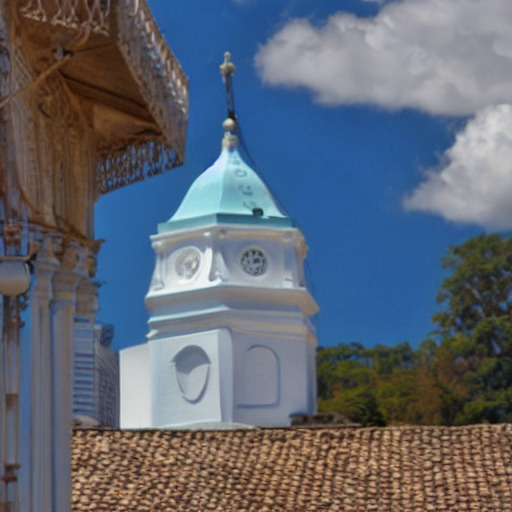}}&
\raisebox{-.5\height}{
\includegraphics[width=0.15\linewidth]{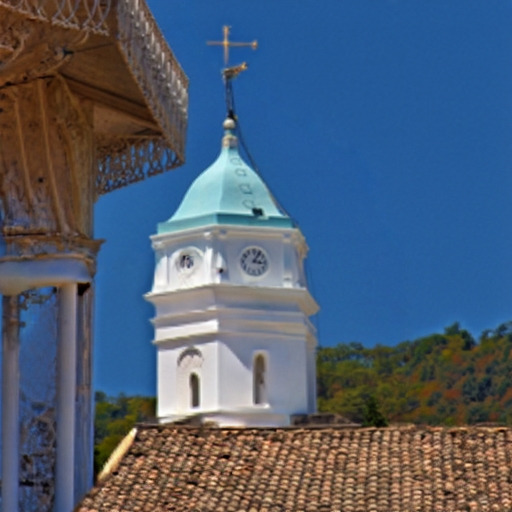}}&
\raisebox{-.5\height}{
\includegraphics[width=0.15\linewidth]{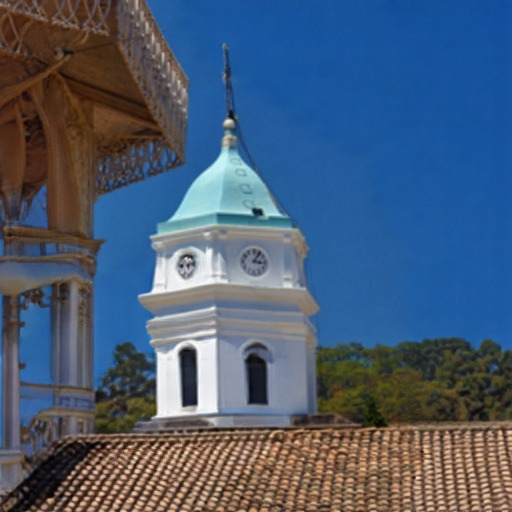}} \\

\raisebox{-.5\height}{
\includegraphics[width=0.15\linewidth]{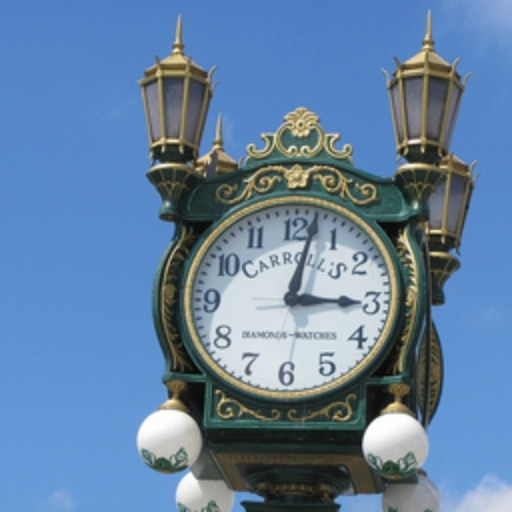}}&
\raisebox{-.5\height}{
\includegraphics[width=0.15\linewidth]{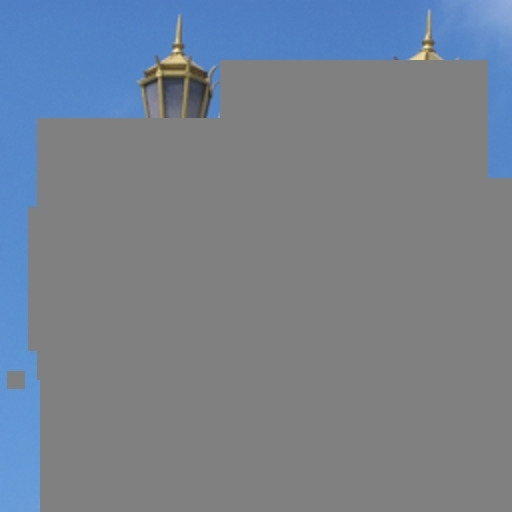}}&
\raisebox{-.5\height}{
\includegraphics[width=0.15\linewidth]{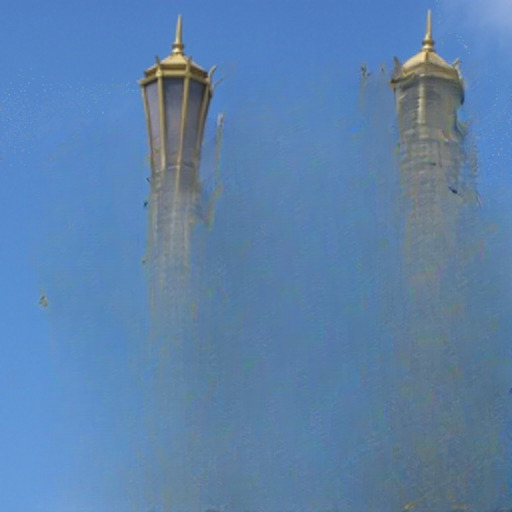}}&
\raisebox{-.5\height}{
\includegraphics[width=0.15\linewidth]{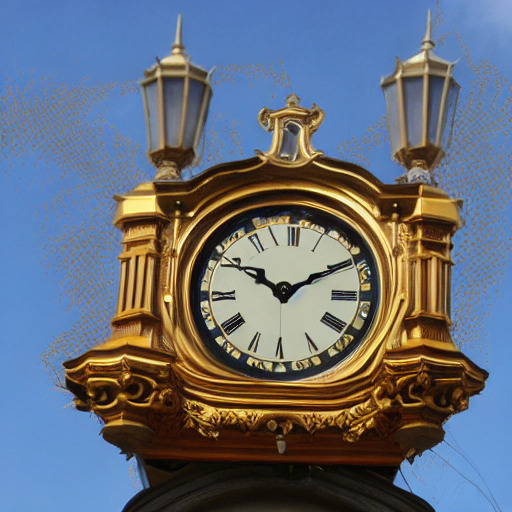}}&
\raisebox{-.5\height}{
\includegraphics[width=0.15\linewidth]{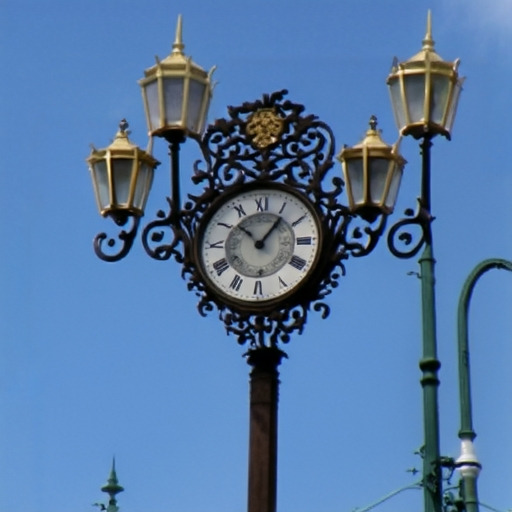}}&
\raisebox{-.5\height}{
\includegraphics[width=0.15\linewidth]{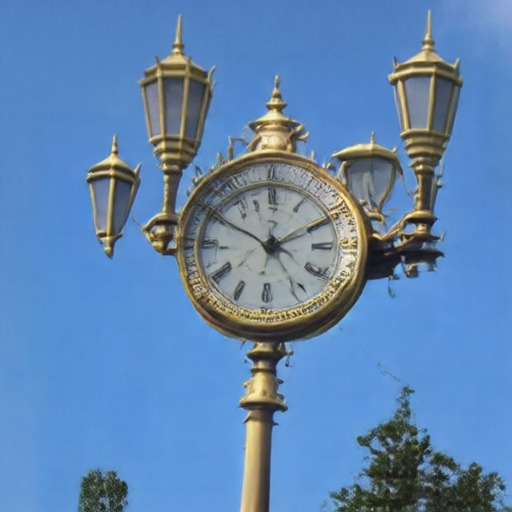}} \\
\end{tabular}
\caption{
\textbf{Qualitative comparison for image inpainting editing.} 
For every masked input image, we compare our distilled SD3-edit inpainting Turbo (1 step) to its teacher SD3-inpainting (50 steps) and other baselines.
}
\label{fig:inpaintcomparison} %
\end{figure*}
}

\newcommand{\teaser}{
\begin{figure}[ht]
\centering
\includegraphics[width=1.0\linewidth]{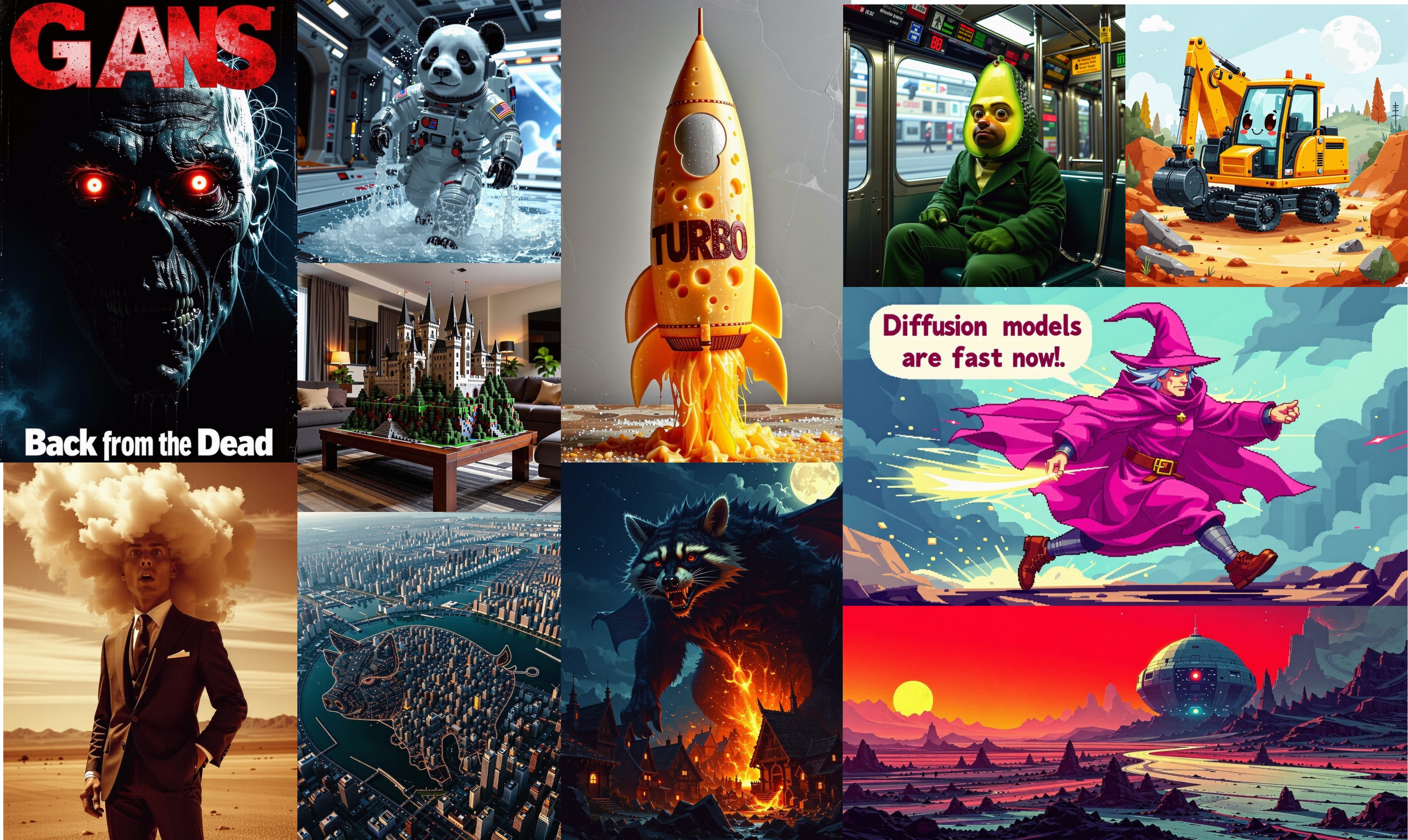}
\captionof{figure}{
\textbf{Generating high-resolution multi-aspect images with \emph{SD3-Turbo}.}
All samples are generated with a maximum of four transformer evaluations trained with latent adversarial diffusion distillation (LADD).
}
\label{fig:teaser}
\end{figure}
}

\newcommand{\system}{
\begin{figure}[t]
\centering
\includegraphics[width=1.0\linewidth]{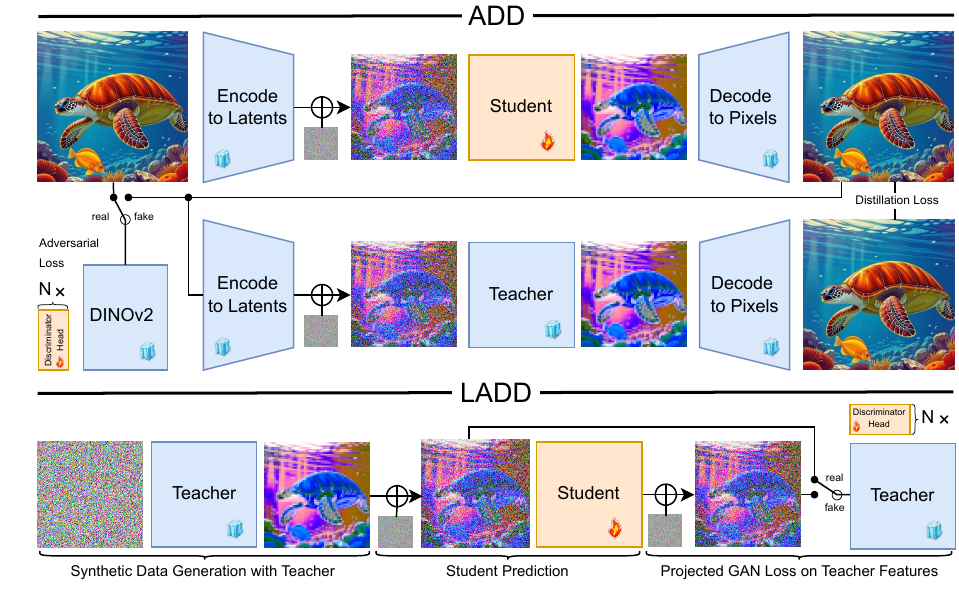}
\captionof{figure}{
\textbf{Comparing ADD and LADD.}
System overview and direct comparison to ADD.
ADD (top two rows) computes a distillation loss in pixel space and an adversarial loss on top of DINOv2 features, thereby requiring expensive decoding from latent space to pixel space. In LADD (bottom row), we use the teacher model for synthetic data generation and its features for the adversarial loss, which allows us to train purely in the latent space.
}
\label{fig:system}
\end{figure}
}

\newcommand{\cherries}{
\begin{figure*}[htp]
    \begin{minipage}[t]{0.325\linewidth}
        \centering
        \includegraphics[width=1.0\linewidth]{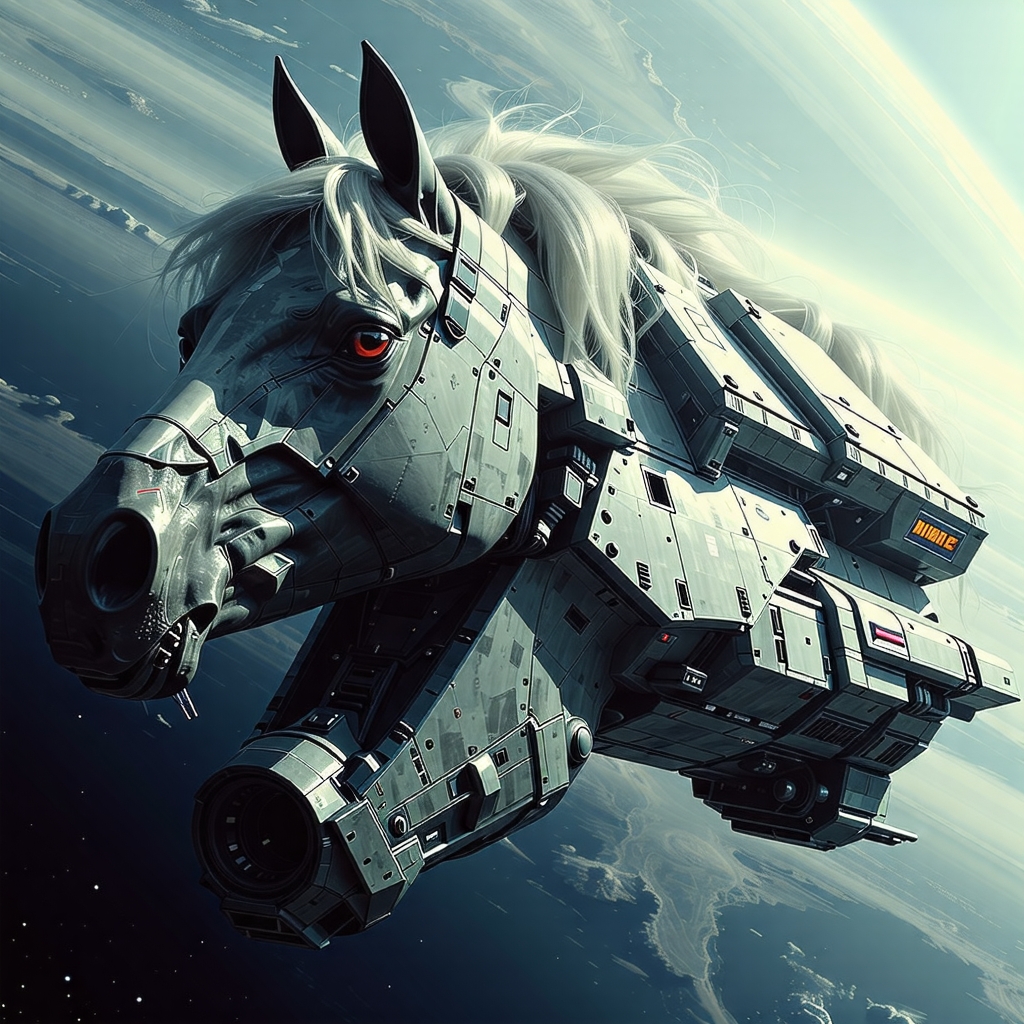} \\
        \tiny{A high-quality photo of a spaceship that looks like the head of a horse.}
    \end{minipage}
    \hfill
    \begin{minipage}[t]{0.325\linewidth}
        \centering
        \includegraphics[width=1.0\linewidth]{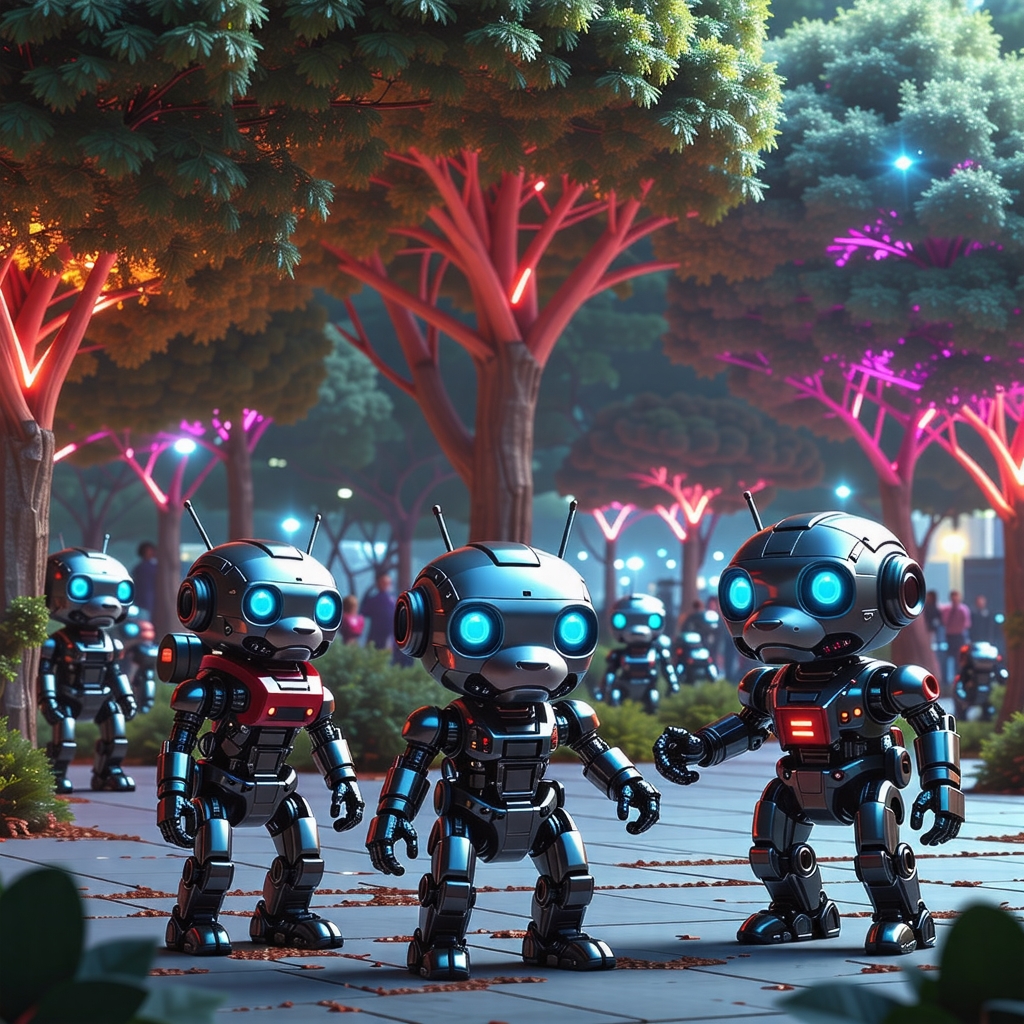} \\
        \tiny{A group of quirky robot animals, with parts made of different metals and machinery, playing in a futuristic park with holographic trees.}
    \end{minipage}
    \hfill
    \begin{minipage}[t]{0.325\linewidth}
        \centering
        \includegraphics[width=1.0\linewidth]{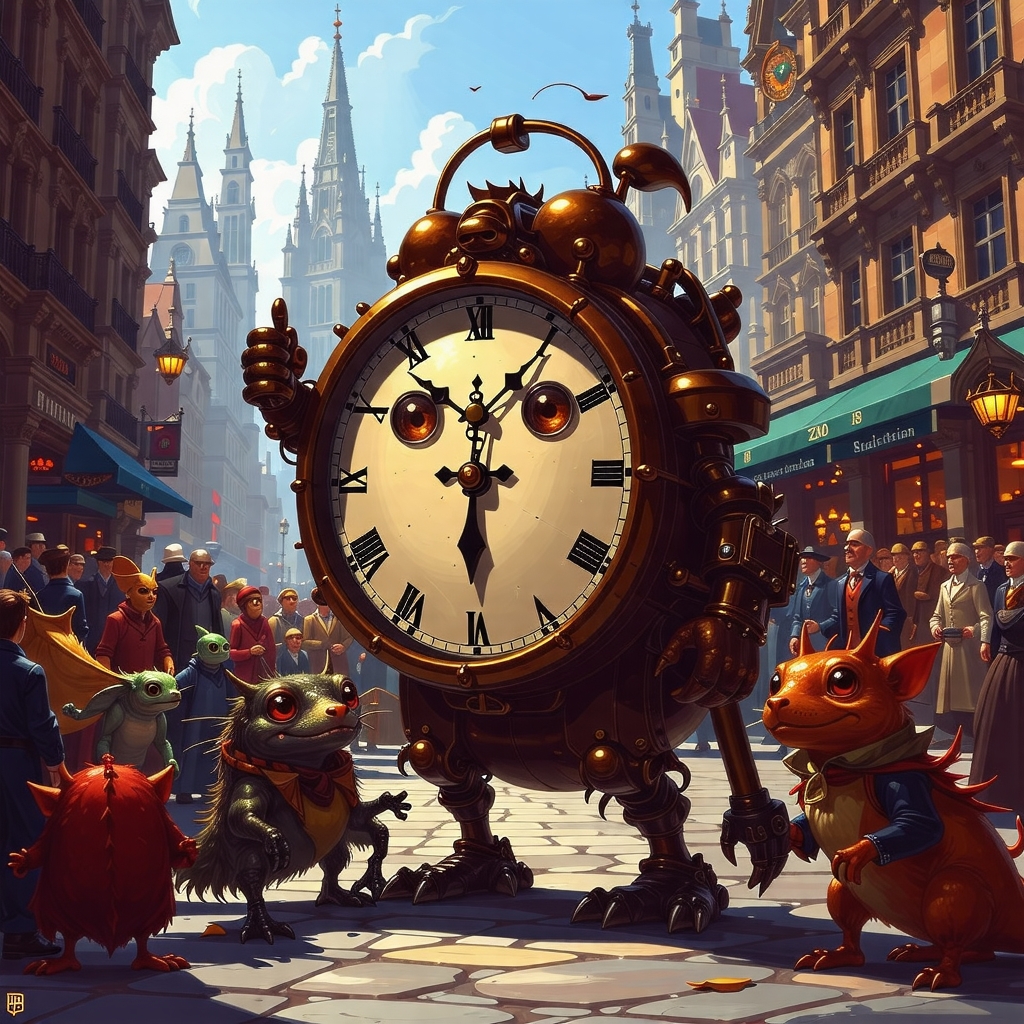} \\
        \tiny{An anthropomorphic clock character in a bustling city square, interacting with time-themed creatures.}
    \end{minipage}
    \hfill
    \vspace{0.5em}
    \centering
    \hspace{0.5em}
    \begin{minipage}[t]{0.1425\linewidth}
        \centering
        \includegraphics[width=1.0\linewidth]{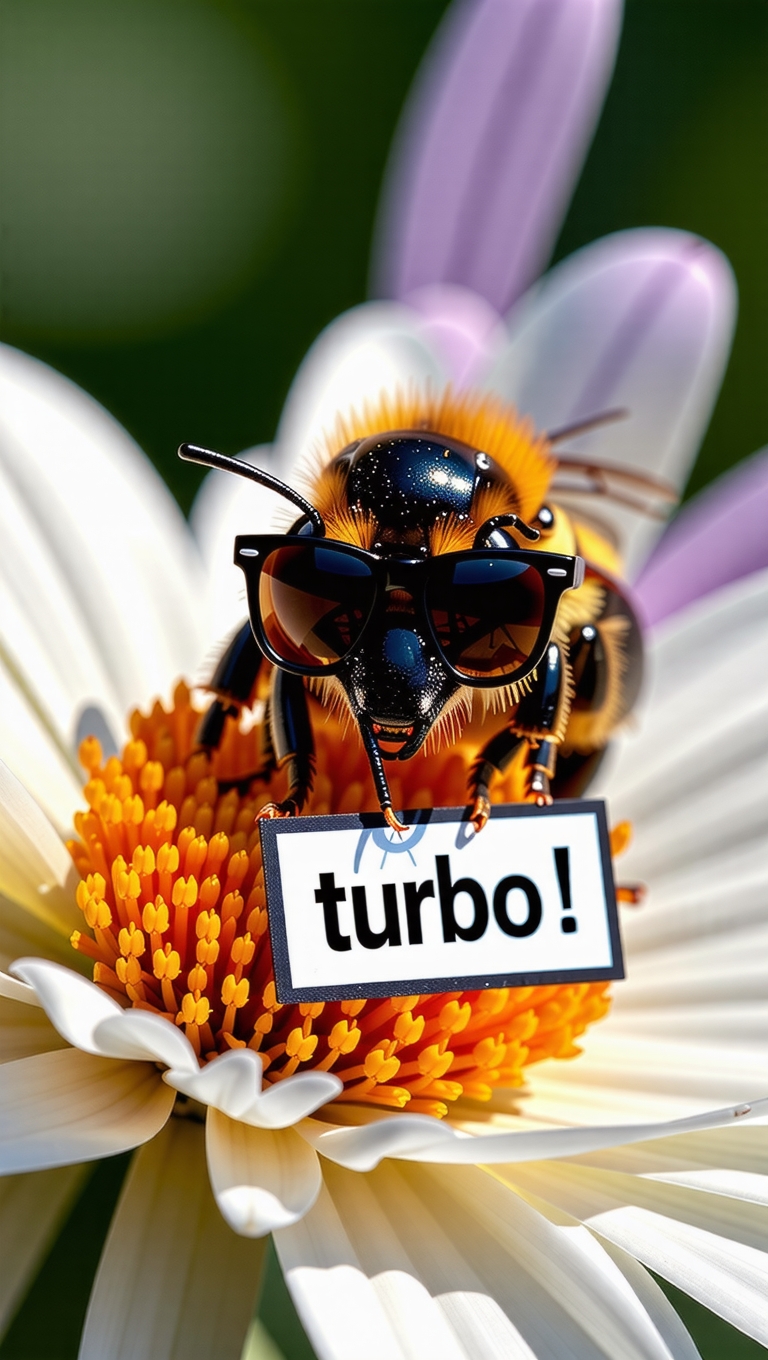} \\
        \tiny{A macro shot of a flower with a bee wearing sunglasses on it that holds a sign saying: "turbo!"}
    \end{minipage}
    \hfill
    \centering
    \begin{minipage}[t]{0.1425\linewidth}
        \centering
        \includegraphics[width=1.0\linewidth]{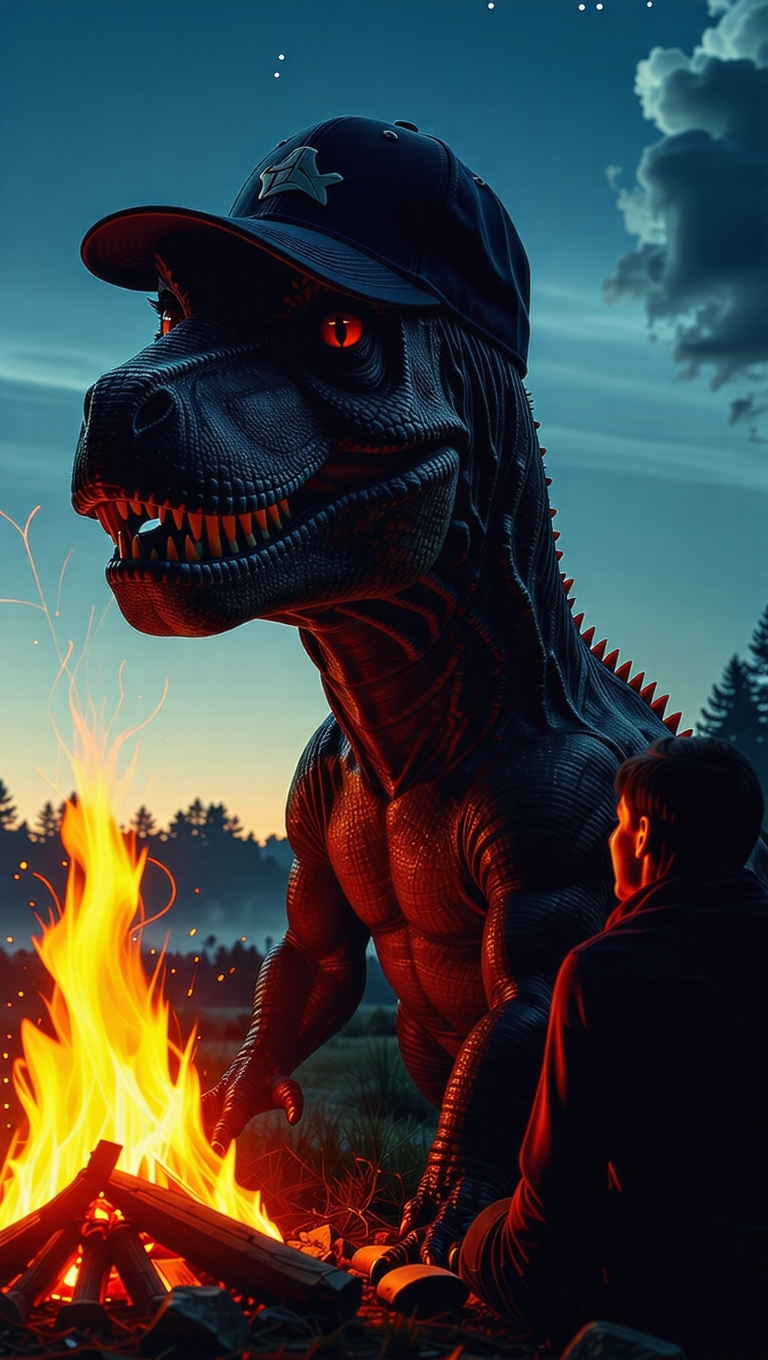} \\
        \tiny{Photo of a T-Rex wearing a cap sitting at a bonfire with his human friend}
    \end{minipage}
    \hfill
    \centering
    \begin{minipage}[t]{0.1425\linewidth}
        \centering
        \includegraphics[width=1.0\linewidth]{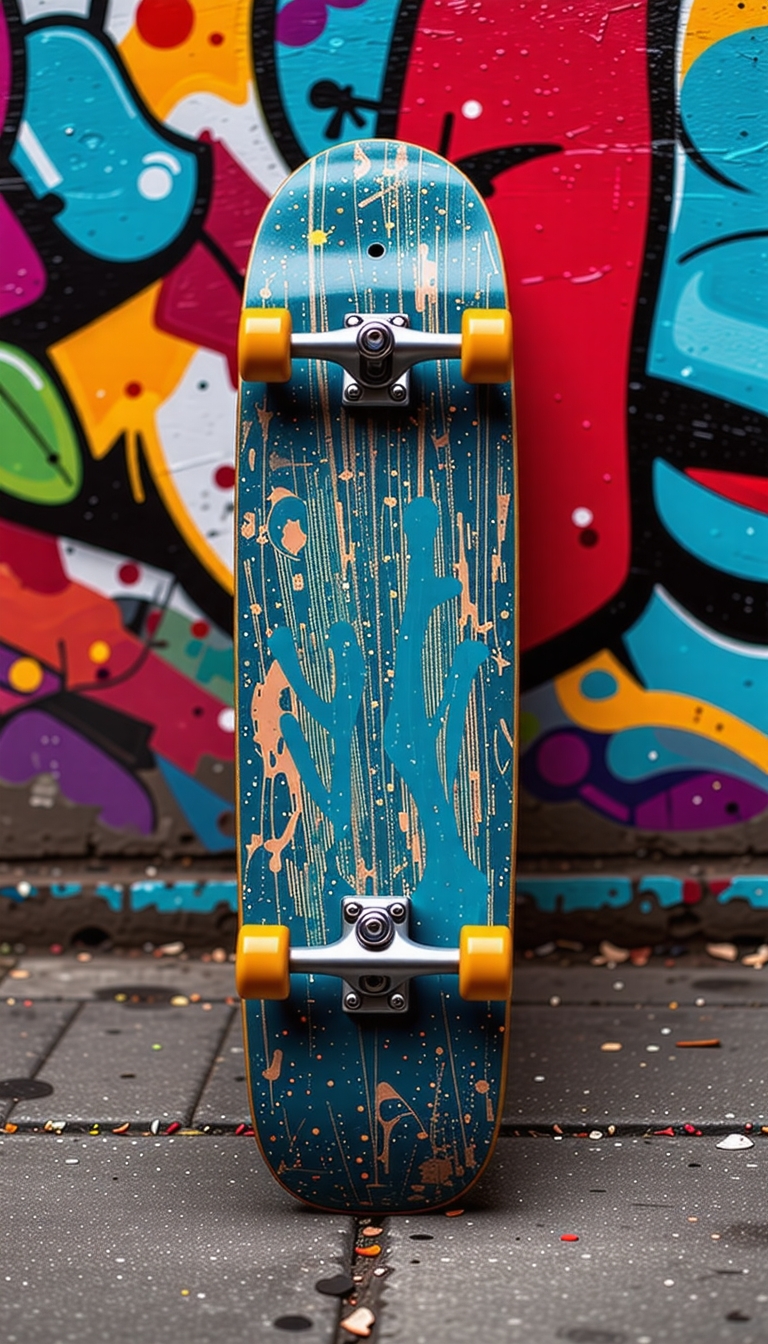} \\
        \tiny{A close-up shot of a skateboard on a colorful graffiti-filled backdrop in an urban setting, capturing the essence of street culture.}
    \end{minipage}
    \hfill
    \centering
    \begin{minipage}[t]{0.1425\linewidth}
        \centering
        \includegraphics[width=1.0\linewidth]{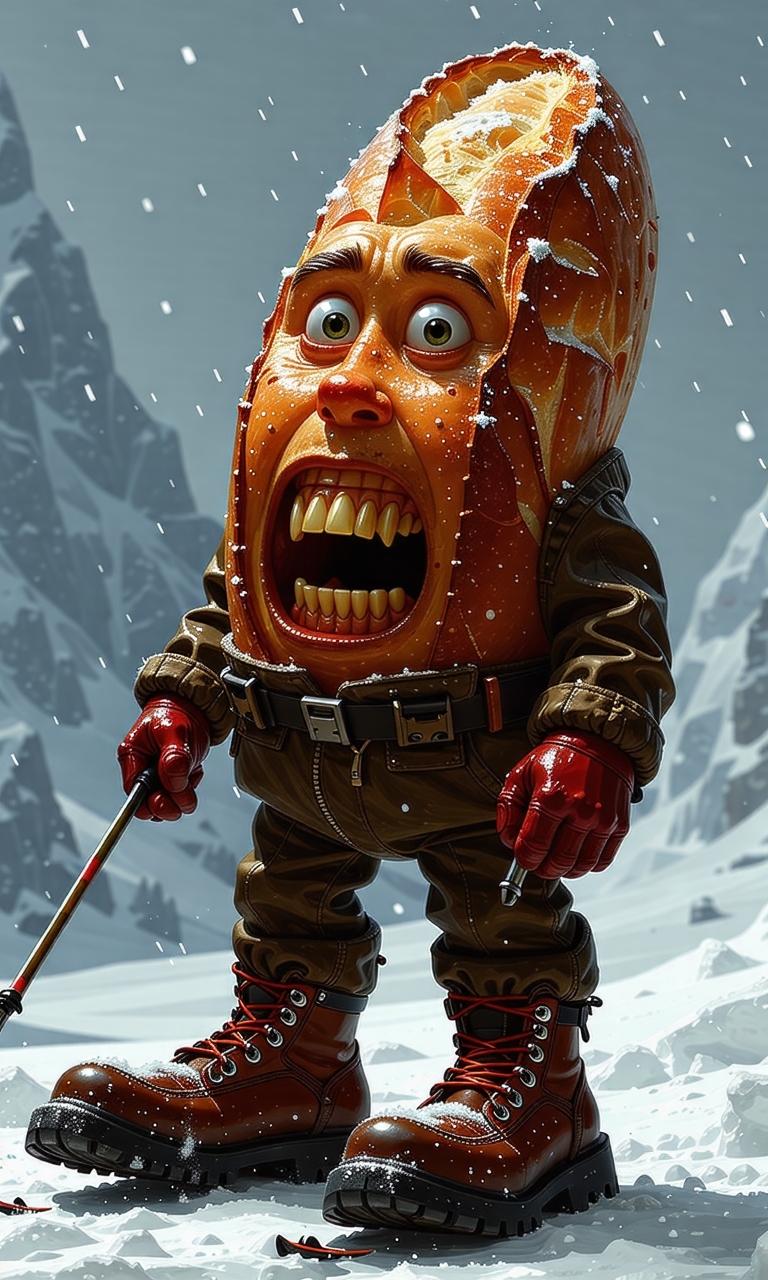} \\
        \tiny{A realistic, detailed photograph of a baguette with human teeth. The baguette is wearing hiking boots and an old-school skiing suit.}
    \end{minipage}
    \hfill
    \centering
    \begin{minipage}[t]{0.1425\linewidth}
        \centering
        \includegraphics[width=1.0\linewidth]{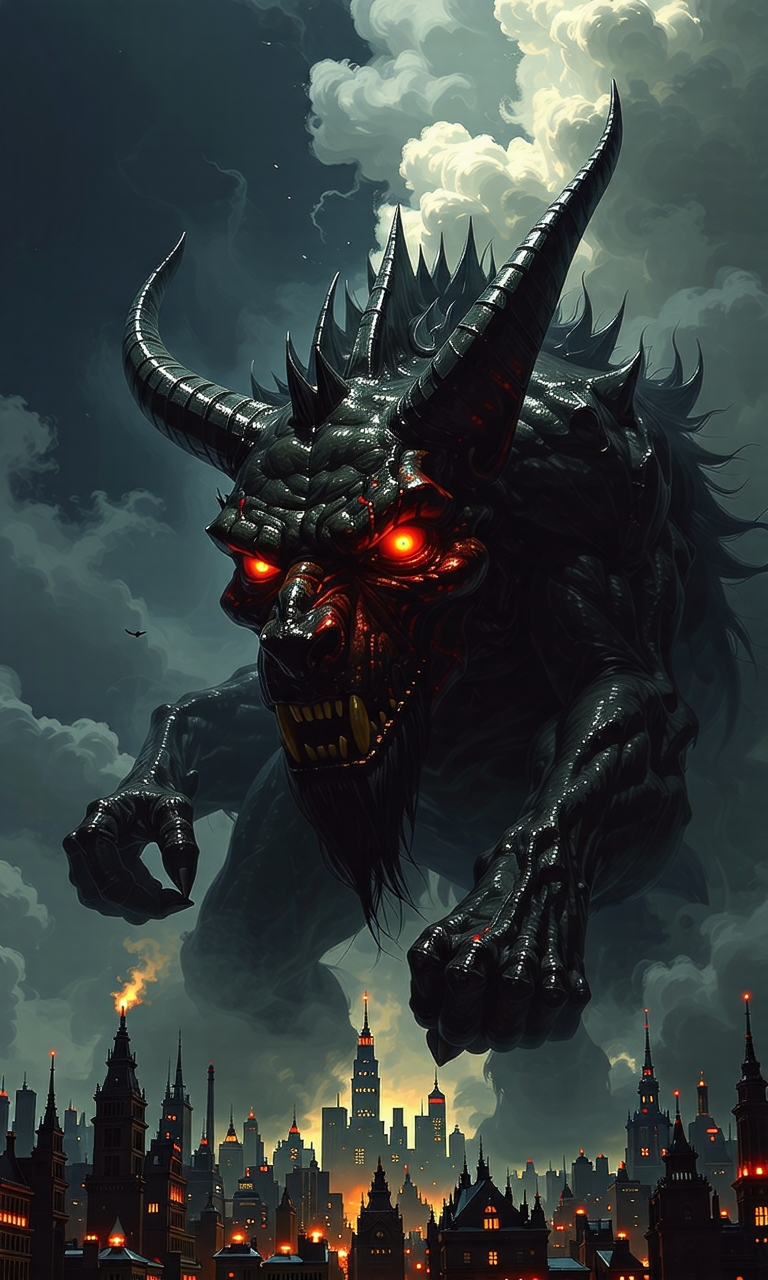} \\
        \tiny{Moloch whose eyes are a thousand blind windows, whose skyscrapers stand in the long streets, whose smoke-stacks and antennae crown the cities!}
    \end{minipage}
    \hfill
    \centering
    \begin{minipage}[t]{0.1425\linewidth}
        \centering
        \includegraphics[width=1.0\linewidth]{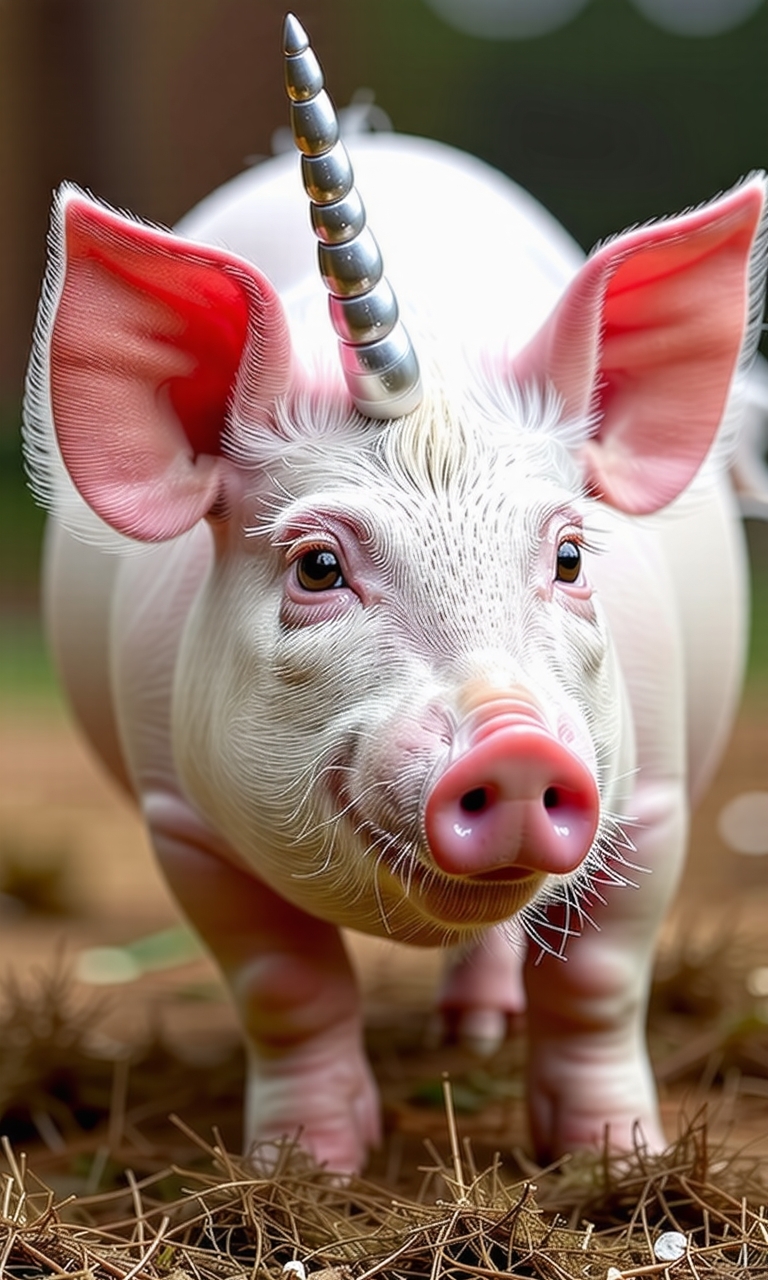} \\
        \tiny{A photograph of a pig with a unicorn's horn.}
    \end{minipage}
    \hfill
    \vspace{0.5em}
    \begin{minipage}[t]{0.495\linewidth}
        \centering
        \includegraphics[width=1.0\linewidth]{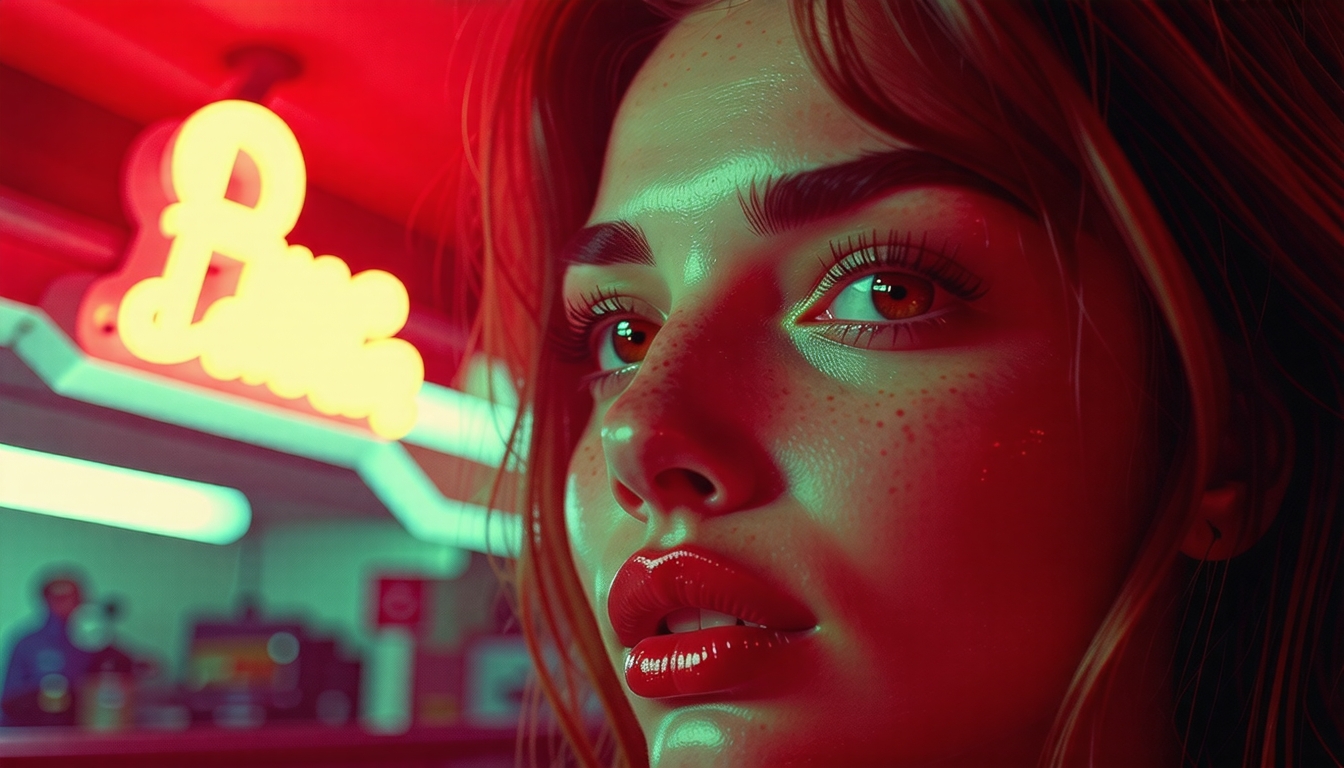} \\
        \tiny{A close-up of a woman's face, lit by the soft glow of a neon sign in a dimly lit, retro diner, hinting at a narrative of longing and nostalgia.}
    \end{minipage}
    \hfill
    \begin{minipage}[t]{0.495\linewidth}
        \centering
        \includegraphics[width=1.0\linewidth]{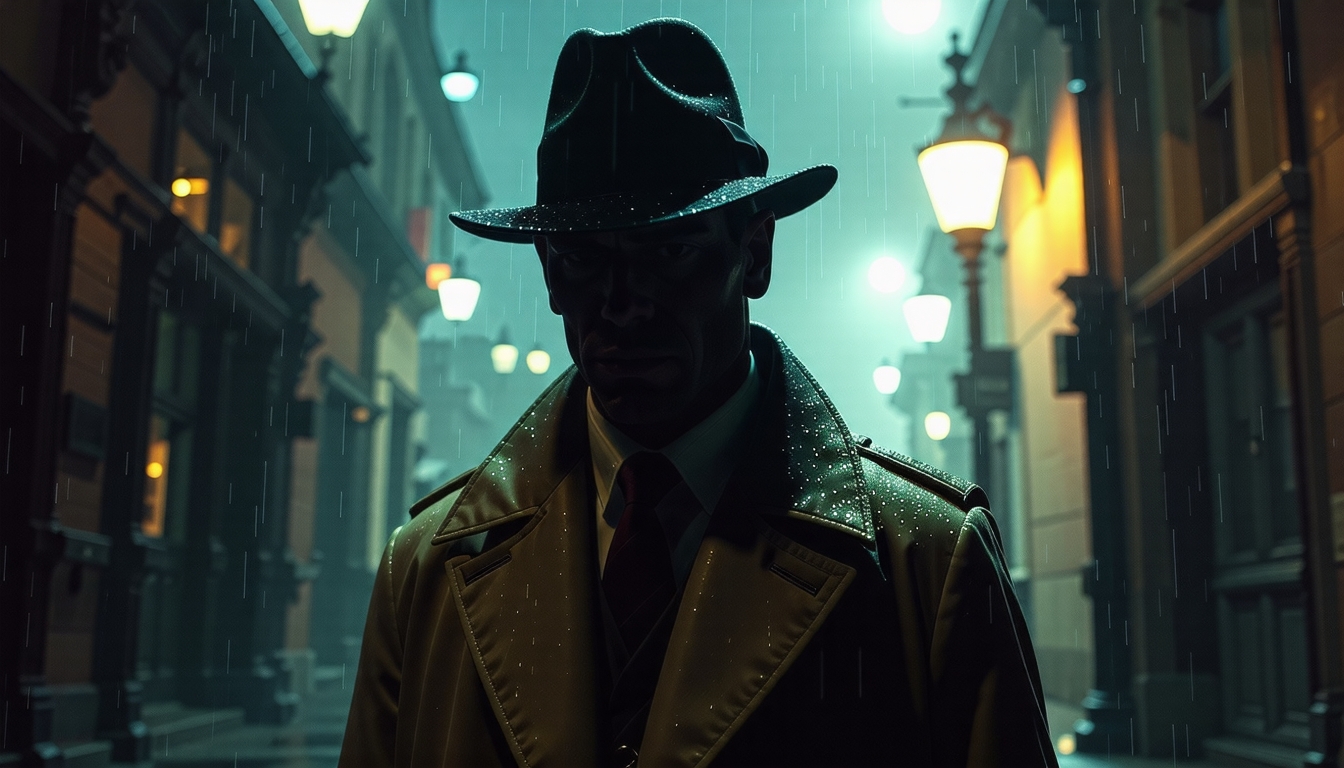} \\
        \tiny{A dramatic shot of a classic detective in a trench coat and fedora, standing in a rain-soaked alleyway under a dim streetlight.
}
    \end{minipage}
    \hfill
    \vspace{0.5em}
    \begin{minipage}[t]{0.49\linewidth}
        \centering
        \includegraphics[width=1.0\linewidth]{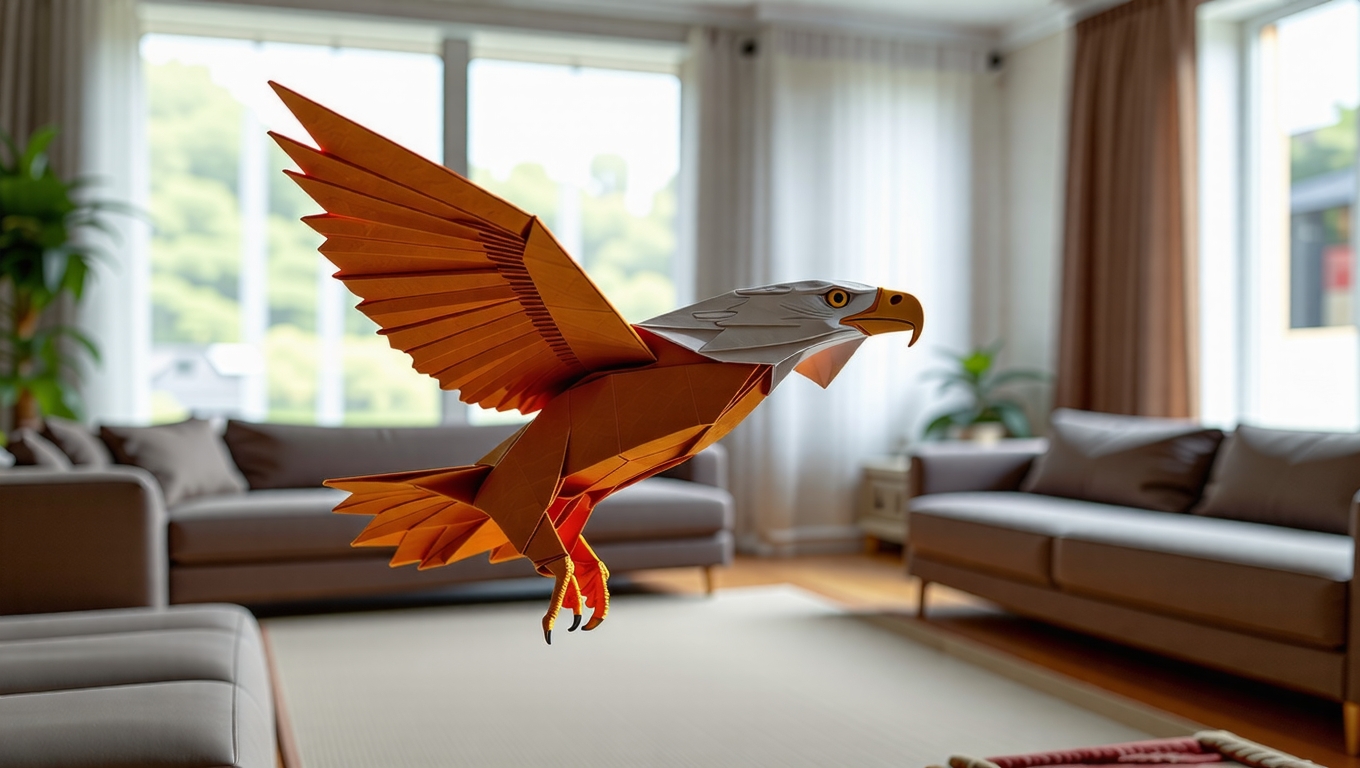} \\
        \tiny{An origami eagle flying through a living room.}
    \end{minipage}
    \hfill
    \begin{minipage}[t]{0.49\linewidth}
        \centering
        \includegraphics[width=1.0\linewidth]{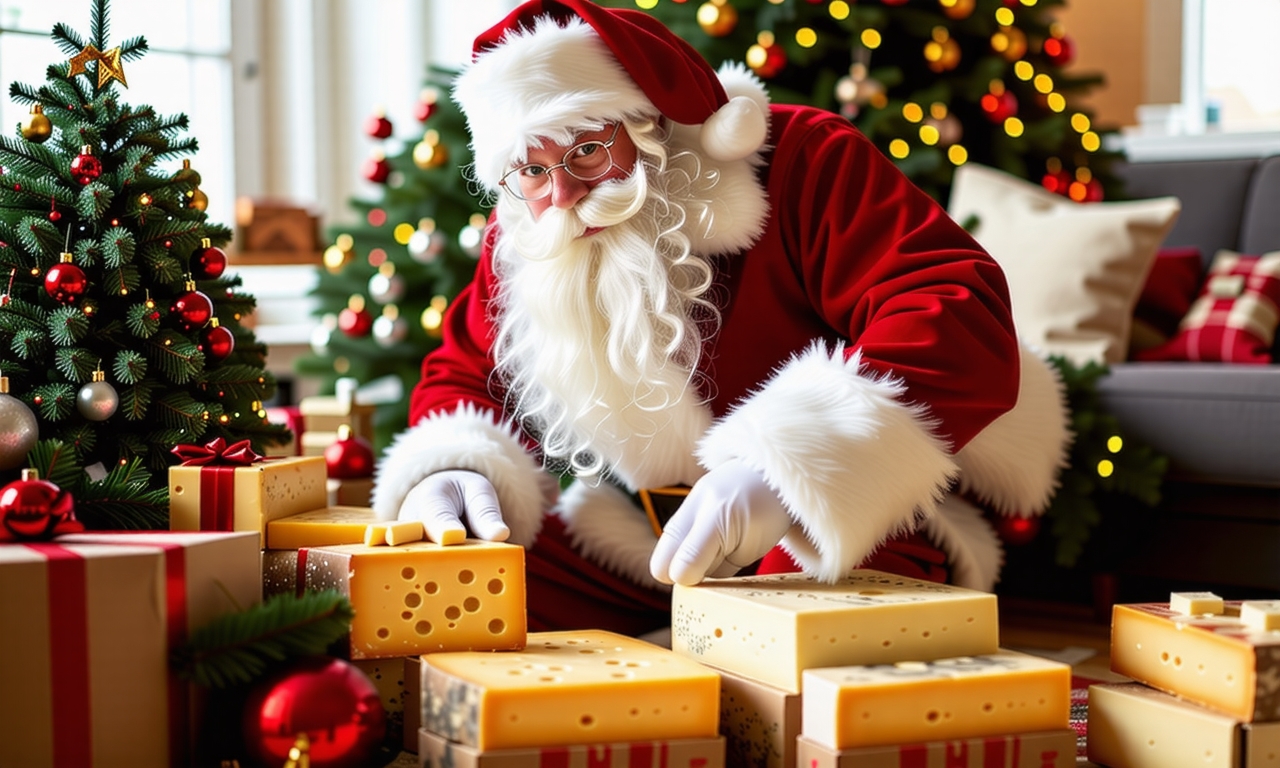} \\
        \tiny{candid photo of santa in my living room placing boxes of cheese under the christmas tree}
    \end{minipage}
\captionof{figure}{
\textbf{\label{fig:cherriesbabycherries}More high-resolution multi-aspect images generated with \emph{SD3-Turbo}.}
All samples are generated with a maximum of four transformer evaluations.
}
\end{figure*}
}

\newcommand{\sigmaschedules}{
\begin{figure*}[t]
\centering
\includegraphics[width=1.0\linewidth]{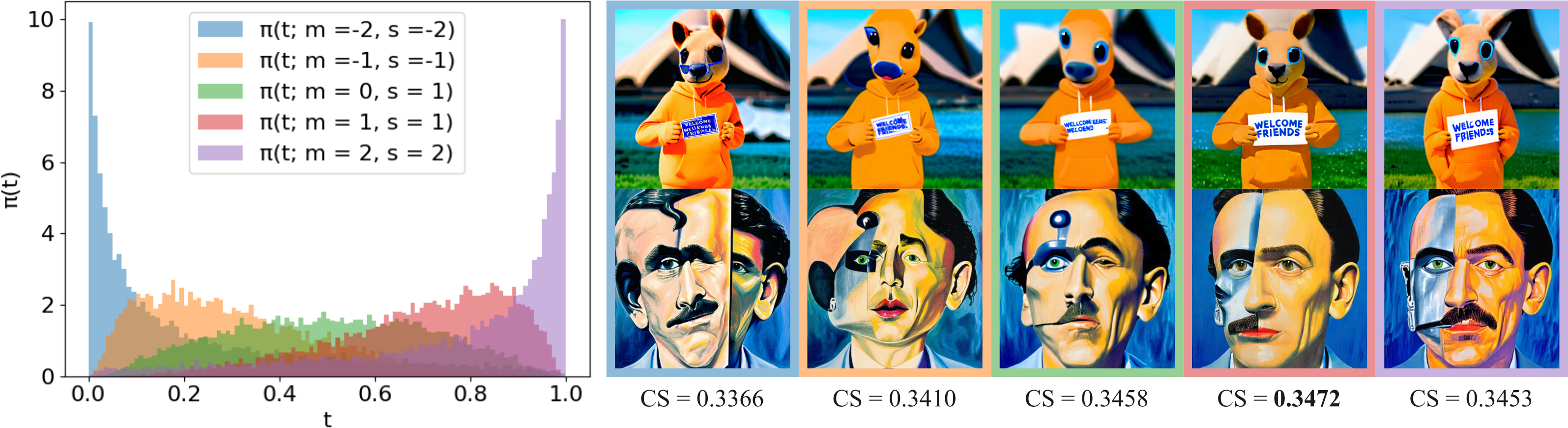}
\captionof{figure}{
\textbf{Controlling the teacher noise distribution.}
We vary the parameters of a logit-normal distribution for biasing the sampling of the teacher noise level.
Shifting to higher noise improves overall coherence. When biasing towards very high noise levels ($m=2, s=2$), we  observe a loss of fine details.
}
\label{fig:sigmaschedules}
\end{figure*}
}

\newcommand{\distillsynthetic}{
\begin{figure*}[t]
\centering
\includegraphics[width=\linewidth]{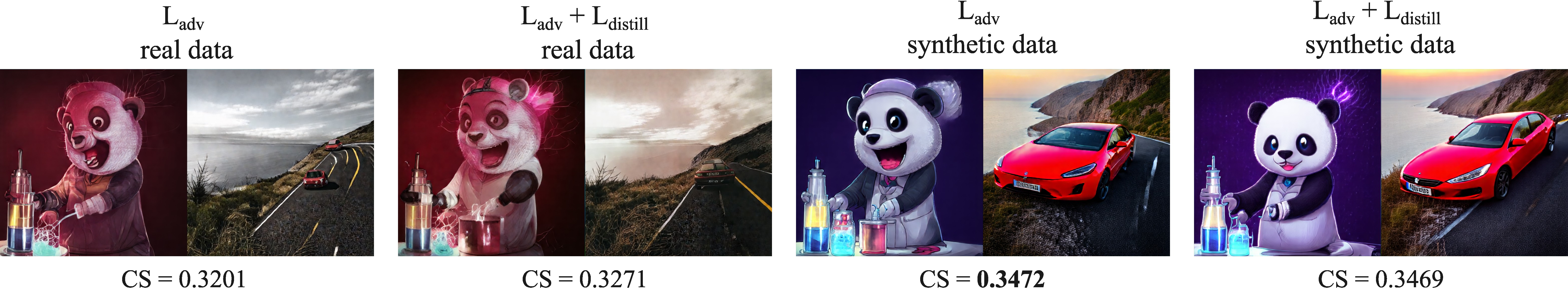}
\captionof{figure}{
\textbf{Synthetic data improves image-text alignment.}
We compare outputs for a fixed seed and the prompts ``panda scientist mixing chemicals" and ``a red car on a scenic road above a cliff."
When training on real data, an additional distillation $L_{distill}$ improves details and thereby increases image-text alignment.
Training on synthetic data substantially outperforms training on real data rendering the distillation loss obsolete.
}
\label{fig:distillsynthetic}
\end{figure*}
}

\newcommand{\lcmvladd}{
\begin{figure*}[t]
\centering
\includegraphics[width=\linewidth]{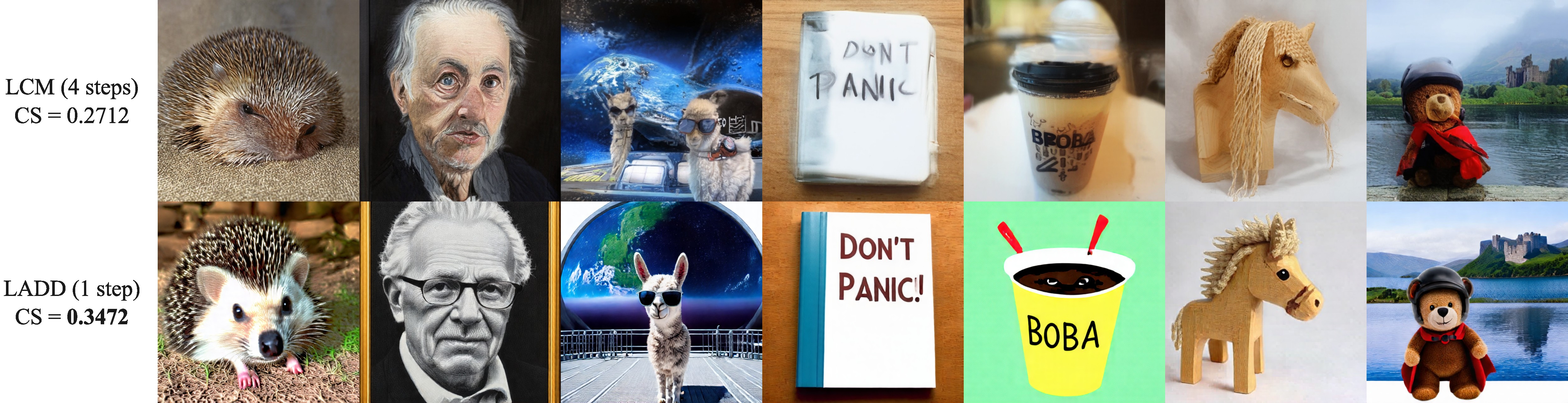}
\captionof{figure}{
\textbf{Comparing latent distillation approaches.}
We distill an MMDiT (depth=24) with both LCM and LADD. For LADD, we use the same model as a teacher and data generator. We find that LADD consistently outperforms LCM in a single step.
}
\label{fig:lcmvladd}
\end{figure*}
}

\newcommand{\dpo}{
\begin{figure*}[t]
\centering
\includegraphics[width=\linewidth]{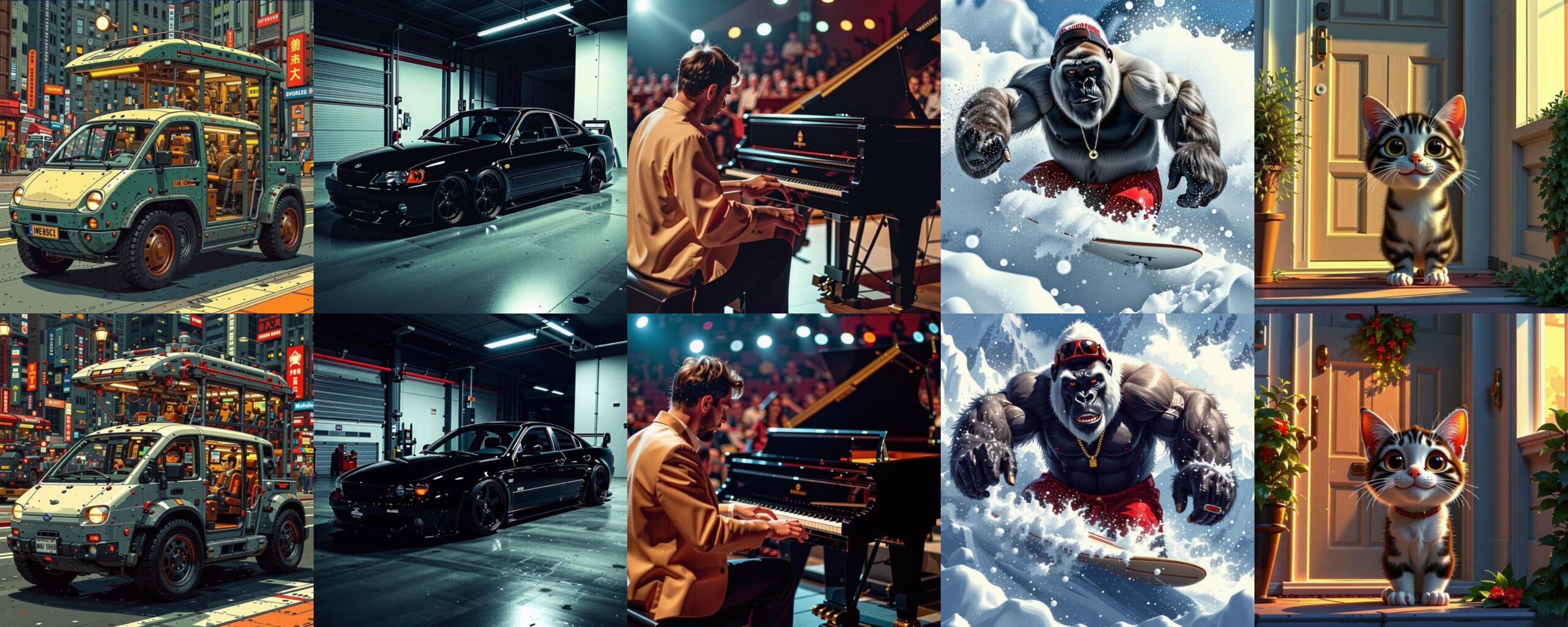}
\captionof{figure}{
\textbf{Applying DPO to LADD students.}
Samples are generated by our best 8B model at 4 steps.
After LADD training, we apply pretrained DPO-LoRA matrices to our student, which adds more details, fixes duplicates objects (e.g. car wheels), improves hands, and increases overall visual appeal (\textit{bottom}).
}
\label{fig:dpo}
\end{figure*}
}

\newcommand{\scaling}{
\begin{figure}[t]
\centering
\includegraphics[width=\linewidth]{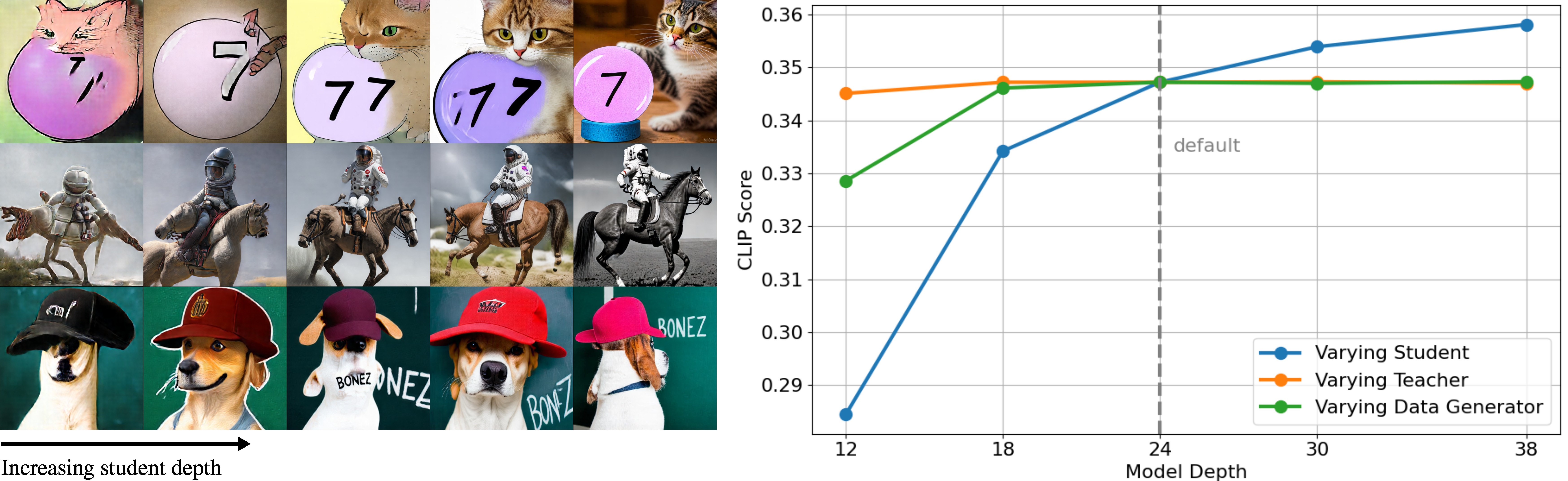}
\captionof{figure}{
\textbf{Scaling behaviour.}
We ablate the size of student, teacher, and data generator model. Our default setting is a depth of 24 for all models and we vary one dimension at a time. A tangible difference is particularly noticeable when varying student depth.
We show samples for a fixed seed and the following prompts:
``a cat patting a crystal ball with the number 7 written on it in black marker", 
``an astronaut riding a horse in a photorealistic style", and
``a dog wearing a baseball cap backwards and writing BONEZ on a chalkboard" (\textit{left, top to bottom}).
}
\label{fig:scaling}
\end{figure}
}

\newcommand{\humanevalallsingle}{
\begin{figure*}[t]
{
\centering
\begin{minipage}{0.49\linewidth}{
\begin{center}
\includegraphics[width=\linewidth]{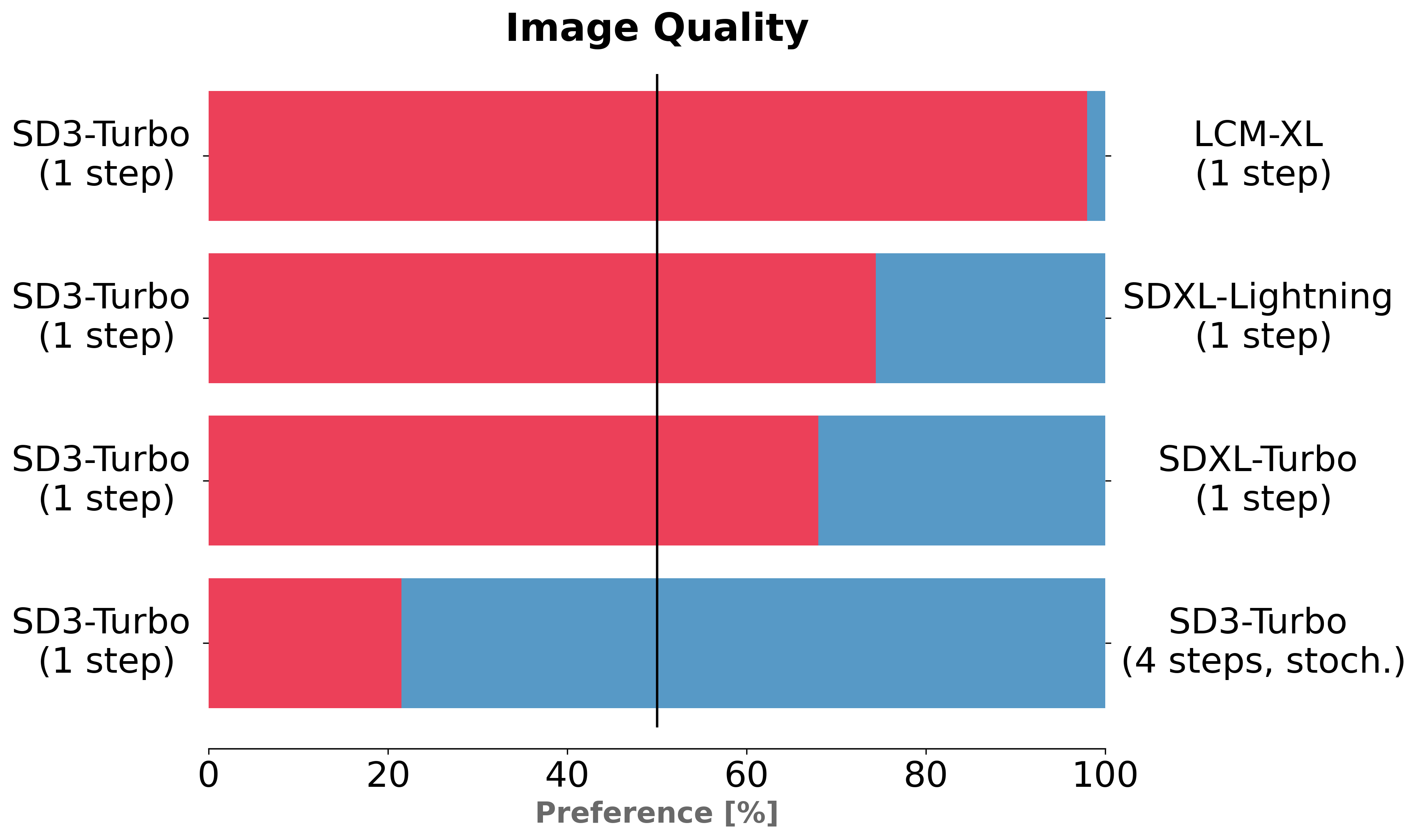}
\end{center}}\end{minipage}
}
\hfill
{
\begin{minipage}{0.49\linewidth}{
\begin{center}
\includegraphics[width=\linewidth]{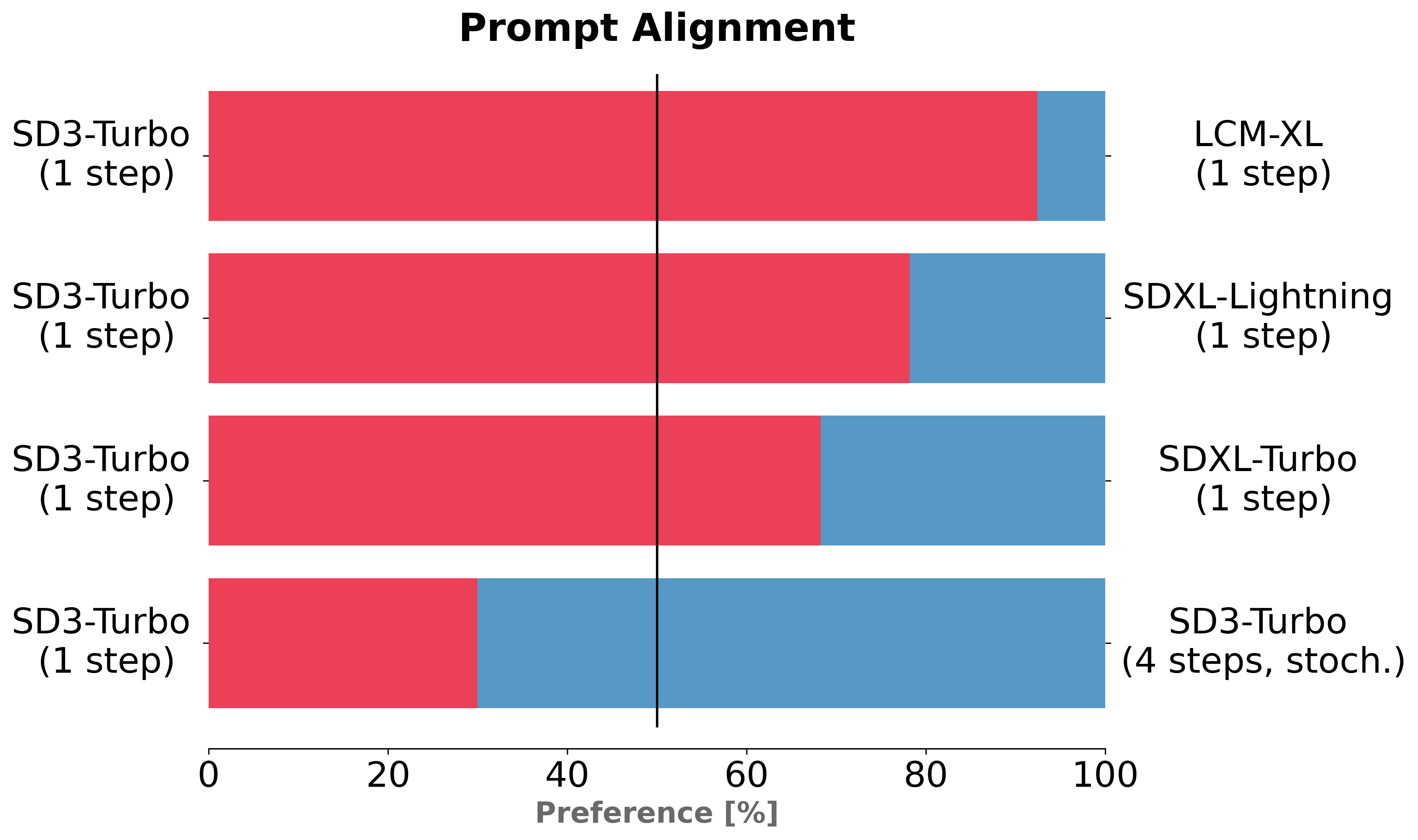}
\end{center}}\end{minipage}
}
\caption{
\textbf{User preference study (\textit{single step}).}
We compare the performance of our model against established baselines.
Our model clearly outperforms all other baselines in human preference for both image quality and prompt alignment.
Using more sampling steps further improves our model's results (bottom row).
}
\label{fig:humanevalallsingle}
\end{figure*}
}

\newcommand{\humanevalallmultiple}{
\begin{figure*}[t]
{
\centering
\begin{minipage}{0.49\linewidth}{
\begin{center}
\includegraphics[width=\linewidth]{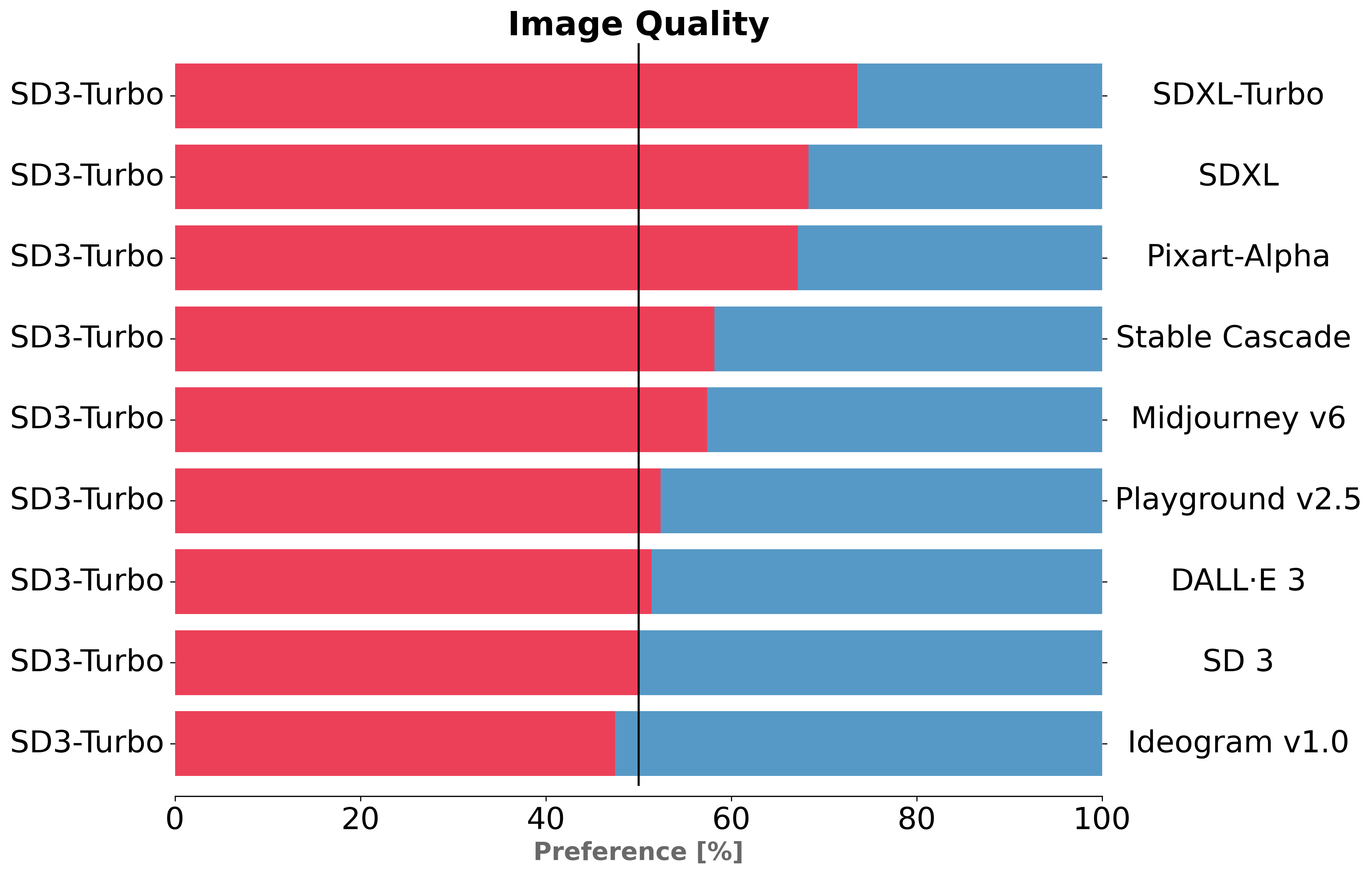}
\end{center}}\end{minipage}
}
\hfill
{
\centering
\begin{minipage}{0.49\linewidth}{
\begin{center}
\includegraphics[width=\linewidth]{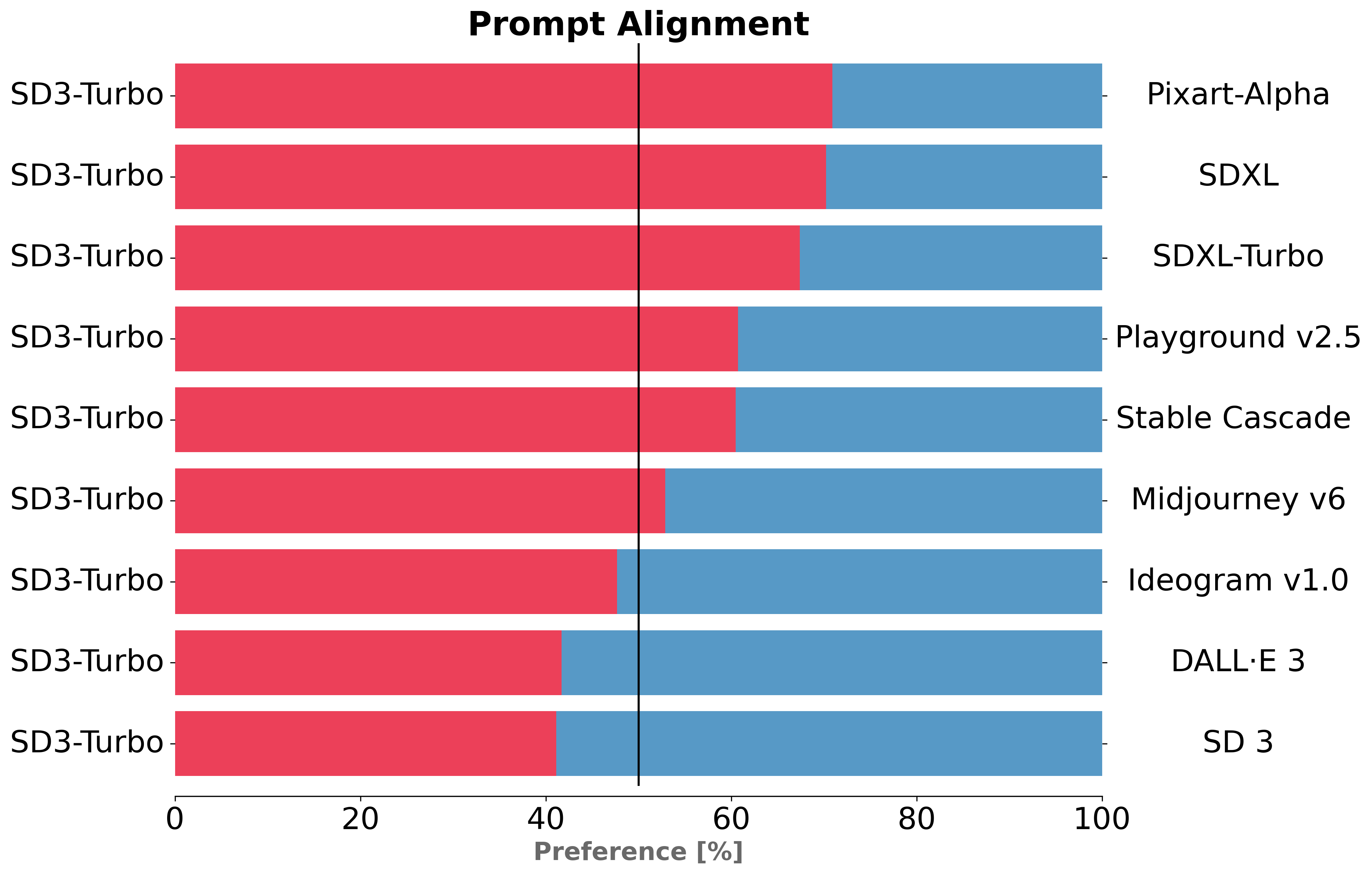}
\end{center}}\end{minipage}
}
\caption{
\textbf{User preference study (\textit{multiple steps}).}
We compare SD3-Turbo $1024^2$-MAR to SOTA text-to-image generators. Our model, using four sampling steps,  outperforms or is on par with all evaluated systems. We use default settings for all other multi-step samplers and four steps for SDXL-Turbo. For the SDXL-Turbo comparison, we downsample the SD3-Turbo outputs to $512^2$ pixels.
}
\label{fig:humanevalallmultiple}
\end{figure*}
}

\newcommand{\imgtoimgmetrics}{
\begin{figure*}[t]
{
\centering
\begin{minipage}{0.49\linewidth}{
\begin{center}
\includegraphics[width=\textwidth]{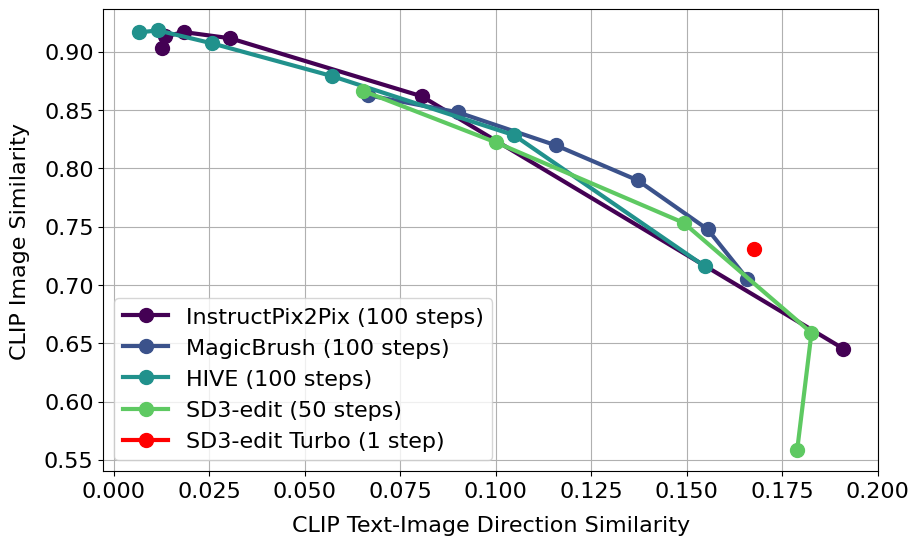}
\end{center}}\end{minipage}
}
\hfill
{
\centering
\begin{minipage}{0.49\linewidth}{
\begin{center}
\begin{tabular}{@{}lcc@{}}
\toprule
                     & FID $\downarrow$ & LPIPS $\downarrow$ \\ \midrule
LaMa                 & 27.21            & \textbf{0.3137}    \\
SD1.5-inpainting     & 10.29            & 0.3879             \\
SD3-inpainting       & \textbf{8.94 }   & 0.3465             \\
SD3-inpainting Turbo & 9.44             & 0.3416             \\ \bottomrule
\end{tabular}
\end{center}
}\end{minipage}
}
\caption{
\textbf{Quantitative evaluation on image-to-image tasks.}
Left: We plot CLIP Image Similarity measuring the  fidelity to the input image over CLIP Direction Similarity measuring the fidelity to the edit prompt; higher is better for both metrics.
We evaluate over varying image conditioning strengths on the PIE-Bench~\citep{ju2023direct} dataset to compare SD3-edit Turbo and baselines. 
Right: Quantitative evaluation of image inpainting on COCO~\citep{lin2014microsoft}; we report FID and LPIPS scores. The masks are created with different policies, ranging from narrow to wide masks and outpainting style masks.
}
\label{fig:imgtoimgmetrics}
\end{figure*}
}

\newcommand{\qualsteps}{
 \begin{figure*}[t]
\centering
\small
\resizebox{\linewidth}{!}{%
\begin{tabular}{llll}
    \vspace{0.5em}
    &\parbox[b]{.3\linewidth}{\tiny\centering \emph{A store front with ’Grassy Meadow’ written on it}}
    &
    \parbox[b]{.3\linewidth}{\tiny\centering \emph{A hot air ballon whose air reservoir is a giant cherry.}}
    &
    \parbox[b]{.3\linewidth}{\tiny \centering \emph{A surreal photograph of a river floating out of an oil painting on a living room wall and spilling over a couch and the wooden floor.$^\dagger$}}
    \\
  \rotatebox[origin=c]{90}{\tiny 1 step} &

  \includegraphics[width=.3\linewidth,valign=m]{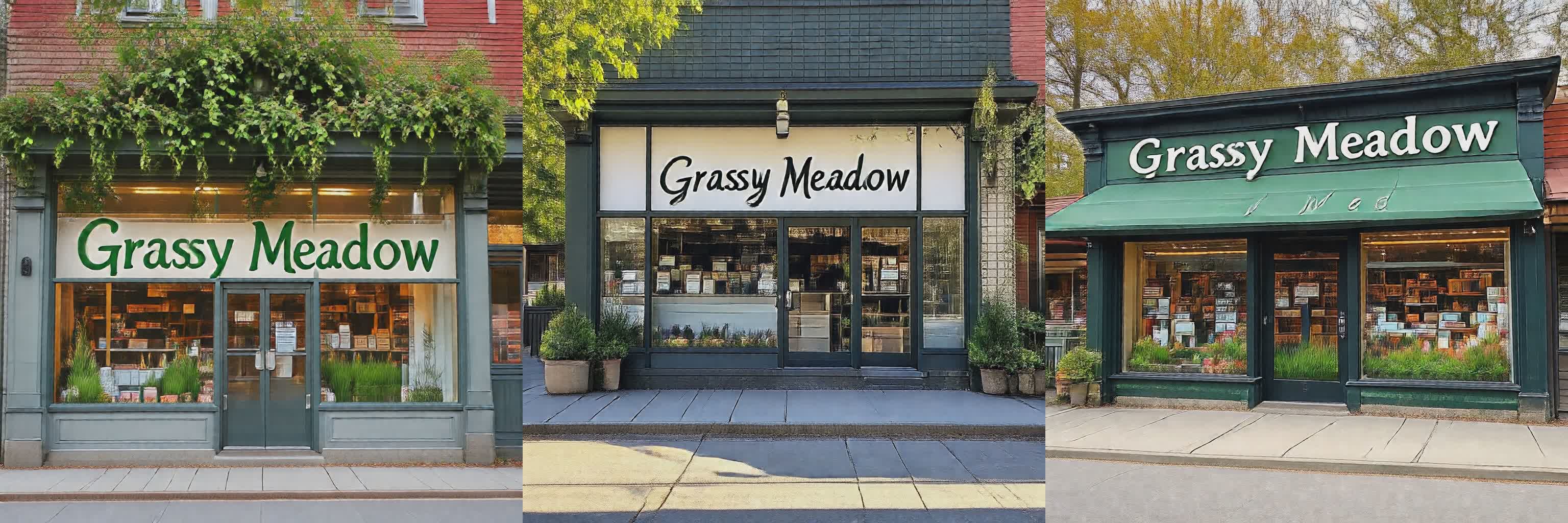}
  &
  \includegraphics[width=.3\linewidth,valign=m]{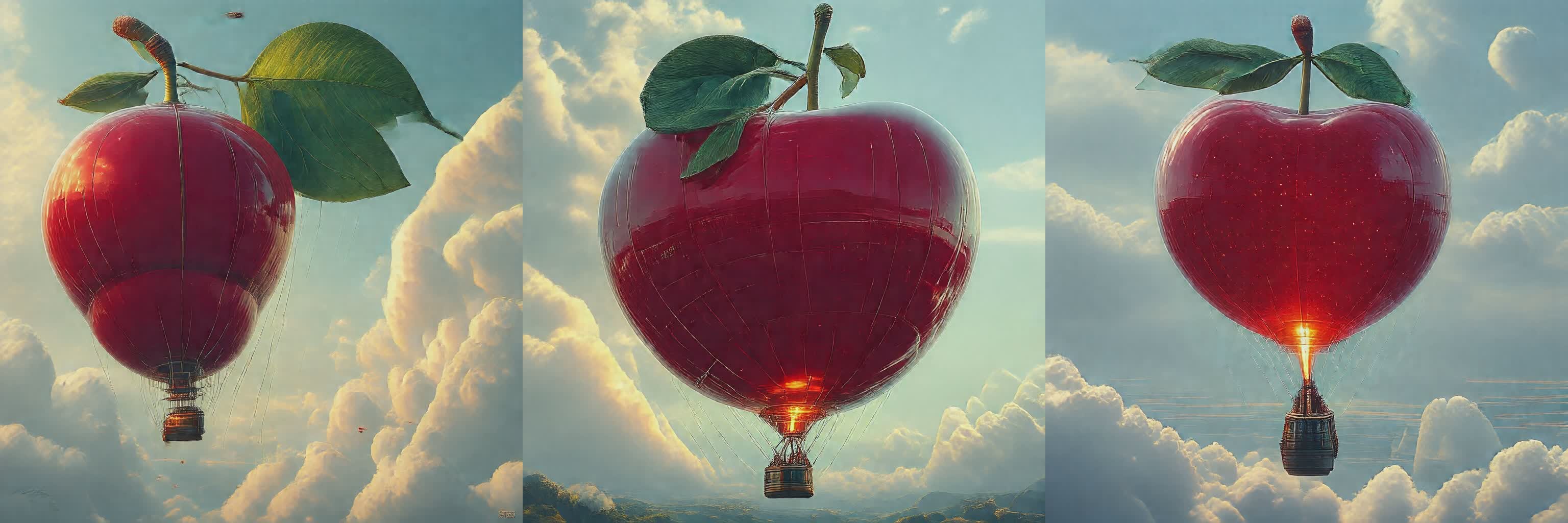}
  &
  \includegraphics[width=.3\linewidth,valign=m]{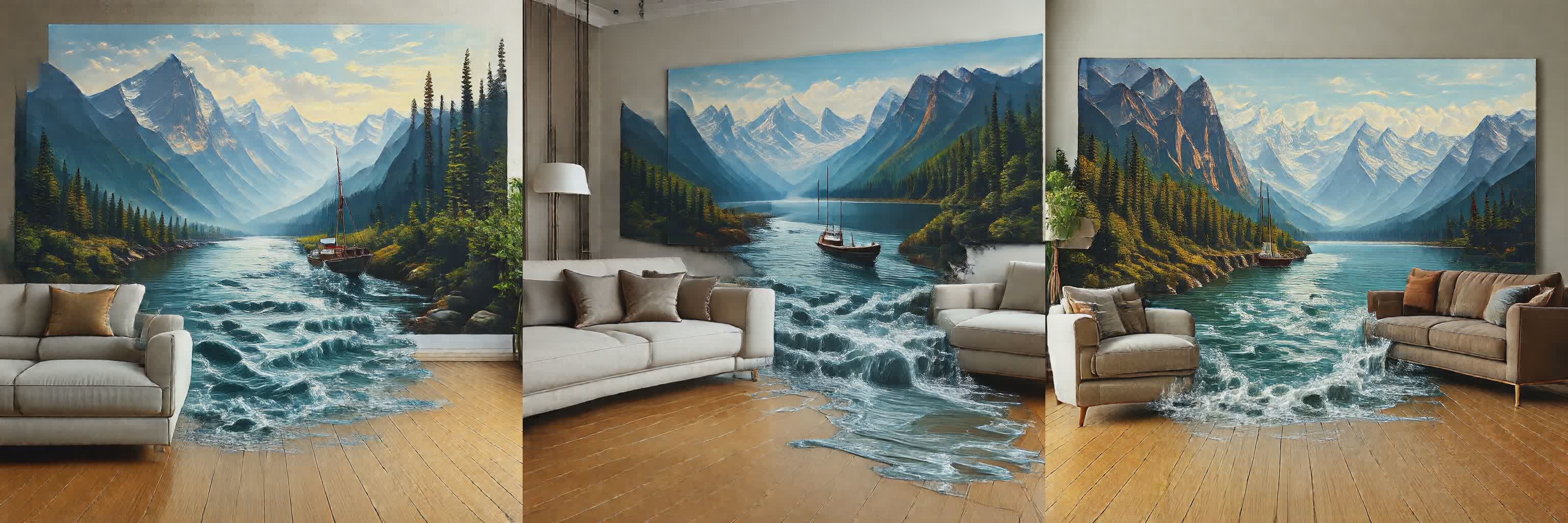}\\
  \rotatebox[origin=c]{90}{\tiny 4 steps} &
  \includegraphics[width=.3\linewidth,valign=m]{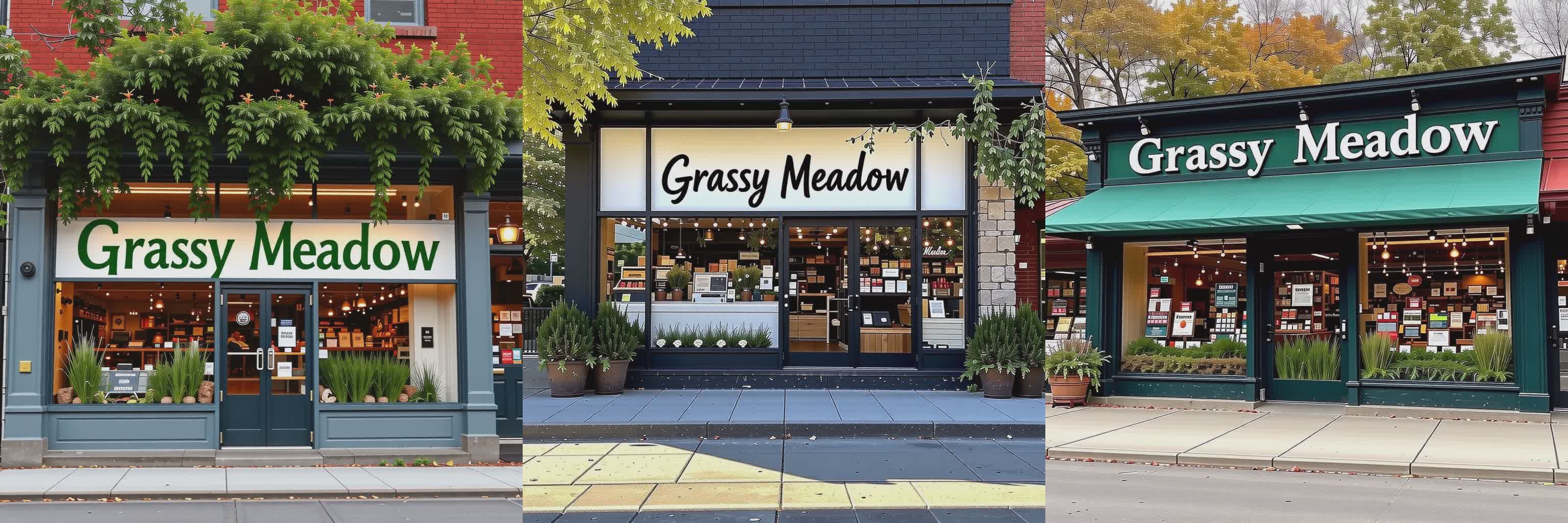}
  &
  \includegraphics[width=.3\linewidth,valign=m]{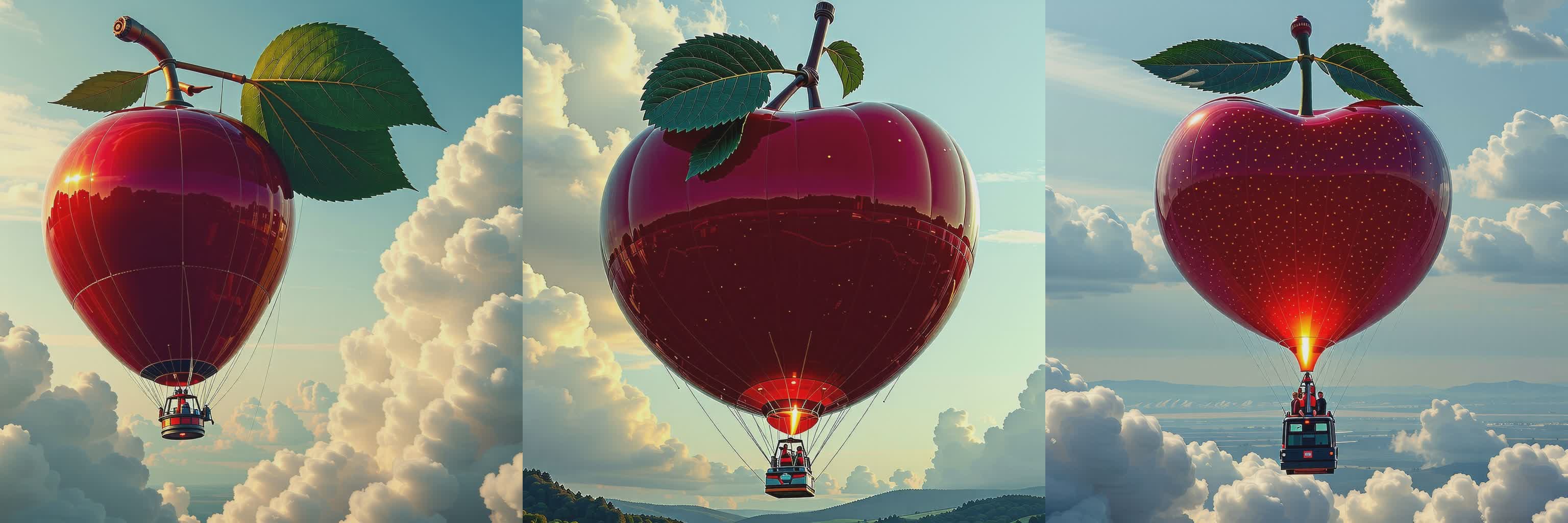}
  &
  \includegraphics[width=.3\linewidth,valign=m]{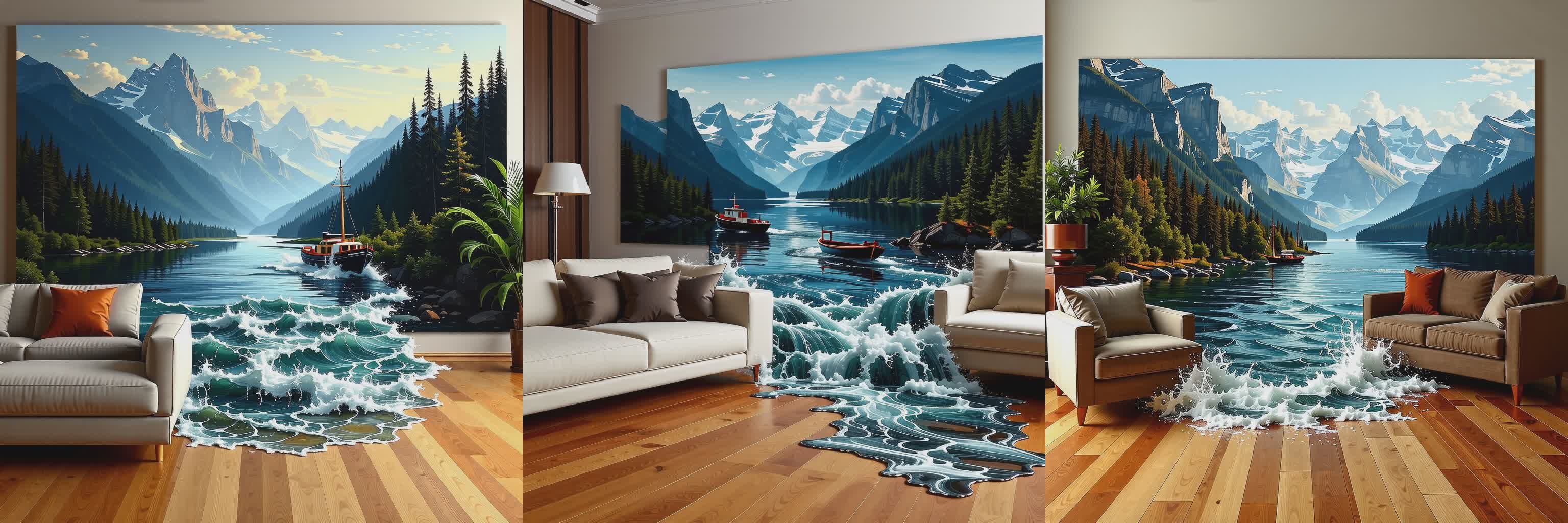}\\
\end{tabular}
}
\caption{\textbf{Qualitative effect of sampling steps.} 
We show qualitative examples when sampling SD3-Turbo with 1 and 4 steps; seeds are constant within columns. $^\dagger$: We only show the first sentence of the prompt to save space. The remainder is as follows: \emph{The painting depicts a tranquil river between mountains. a ship gently bobbing in the water and entering the living room. The river's edge spills onto the wooden floor, merging the world of art with reality. The living room is adorned with tasteful furniture and a warm, inviting atmosphere., cinematic, photo, poster.}.
\label{fig:qualitativesteps}
}
  \end{figure*}

}

\newcommand{\texttoimfailures}{
\begin{figure*}[t]
\centering
\scriptsize
\resizebox{\linewidth}{!}{%
\begin{tabular}{
c@{\hspace{0\tabcolsep}}c
c@{\hspace{0\tabcolsep}}c
c@{\hspace{0\tabcolsep}}c
}
\multicolumn{2}{c}{
\parbox[b]{.29\linewidth}{\tiny\centering \emph{A black dog sitting on a wooden chair. A white cat with black ears is standing up with its paws on the chair.}}
}&
\multicolumn{2}{c}{
\parbox[b]{.29\linewidth}{\tiny\centering \emph{A set of 2x2 emoji icons with happy, angry, surprised and sobbing faces. The emoji icons look like dogs. All of the dogs are wearing blue turtlenecks.}}
}&
\multicolumn{2}{c}{
\parbox[b]{.29\linewidth}{\tiny\centering \emph{a subway train with no cows in it.}}
}\vspace{0.3em}\\
\includegraphics[width=0.15\linewidth]{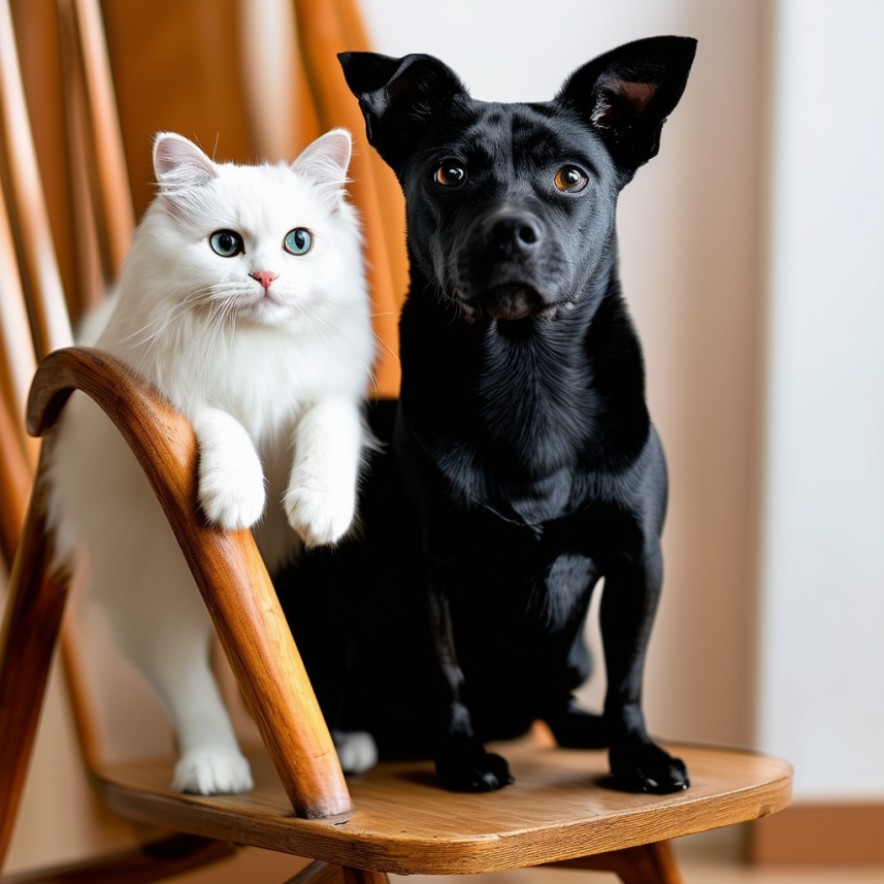}&
\includegraphics[width=0.15\linewidth]{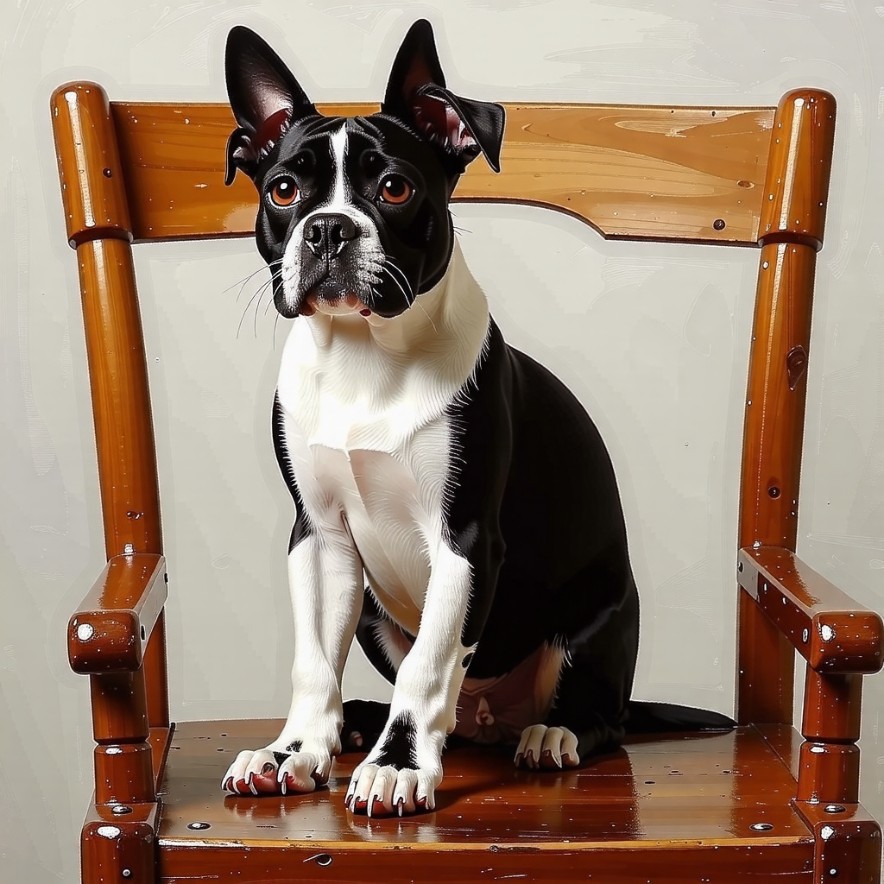}&
\includegraphics[width=0.15\linewidth]{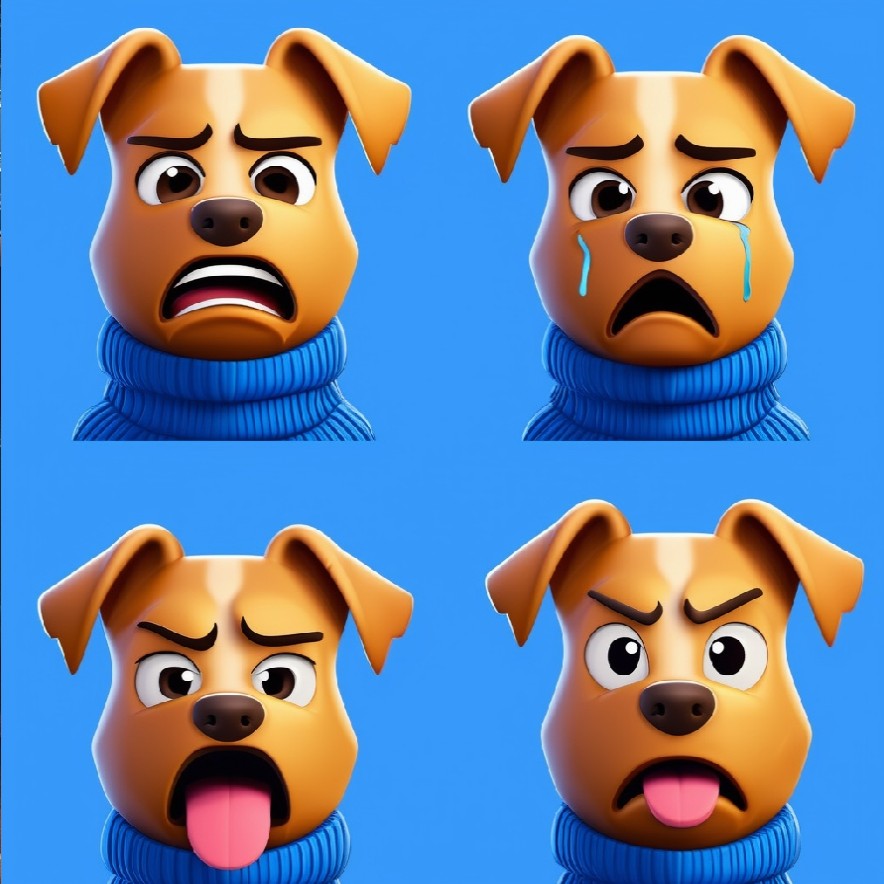}&
\includegraphics[width=0.15\linewidth]{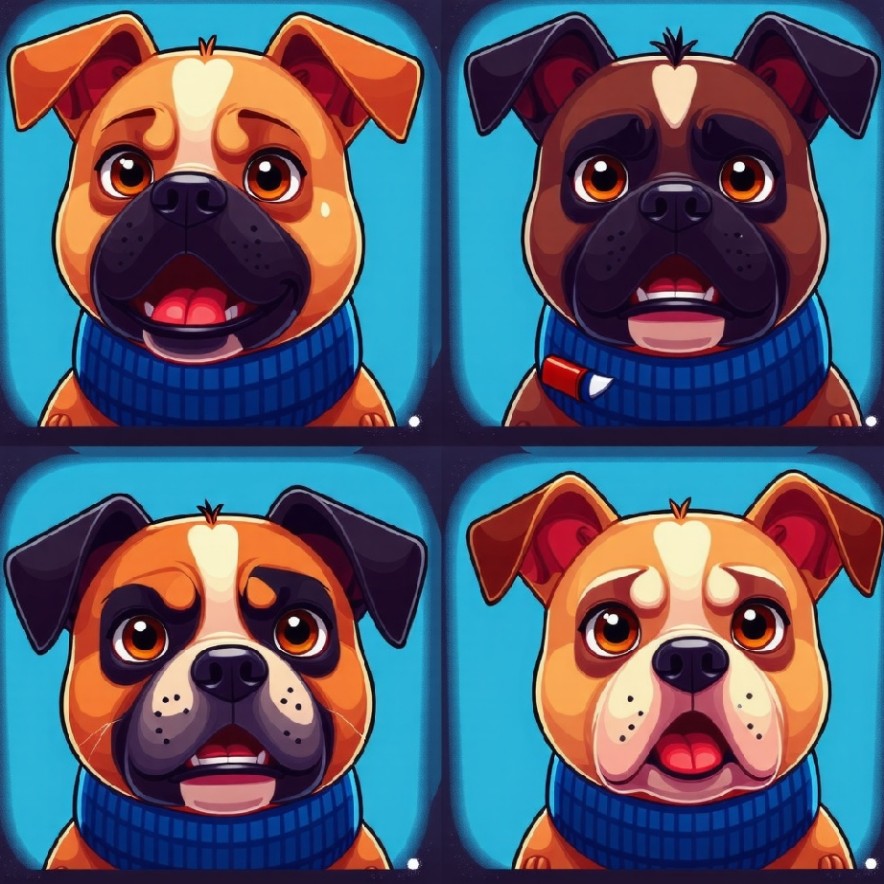}&
\includegraphics[width=0.15\linewidth]{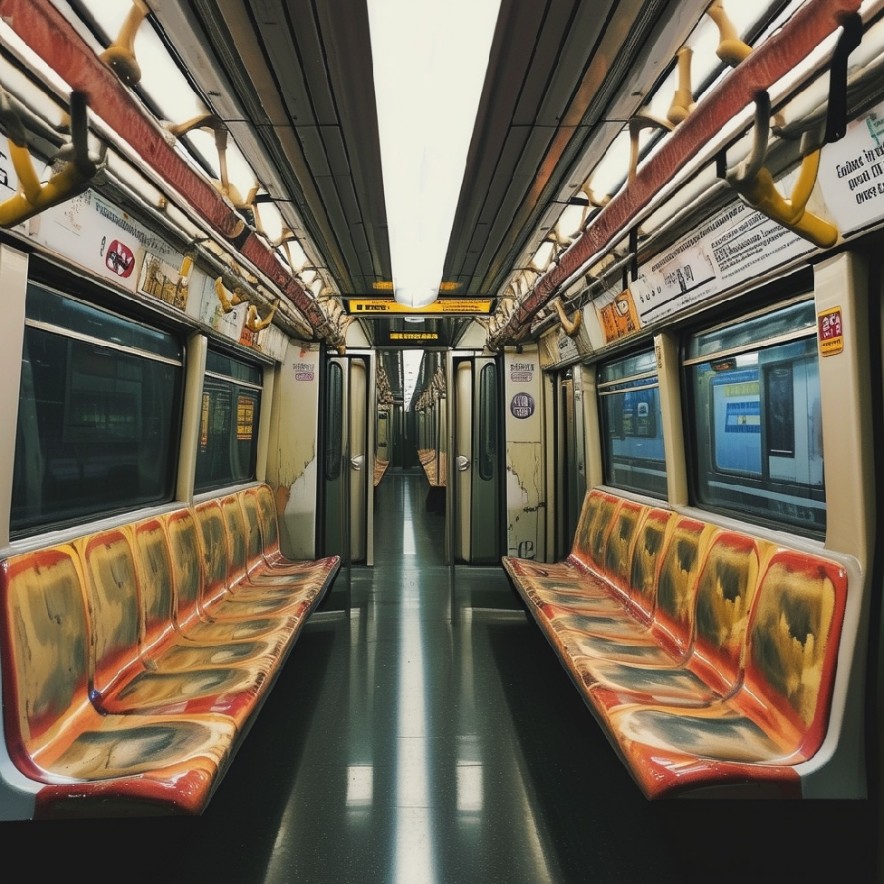}&
\includegraphics[width=0.15\linewidth]{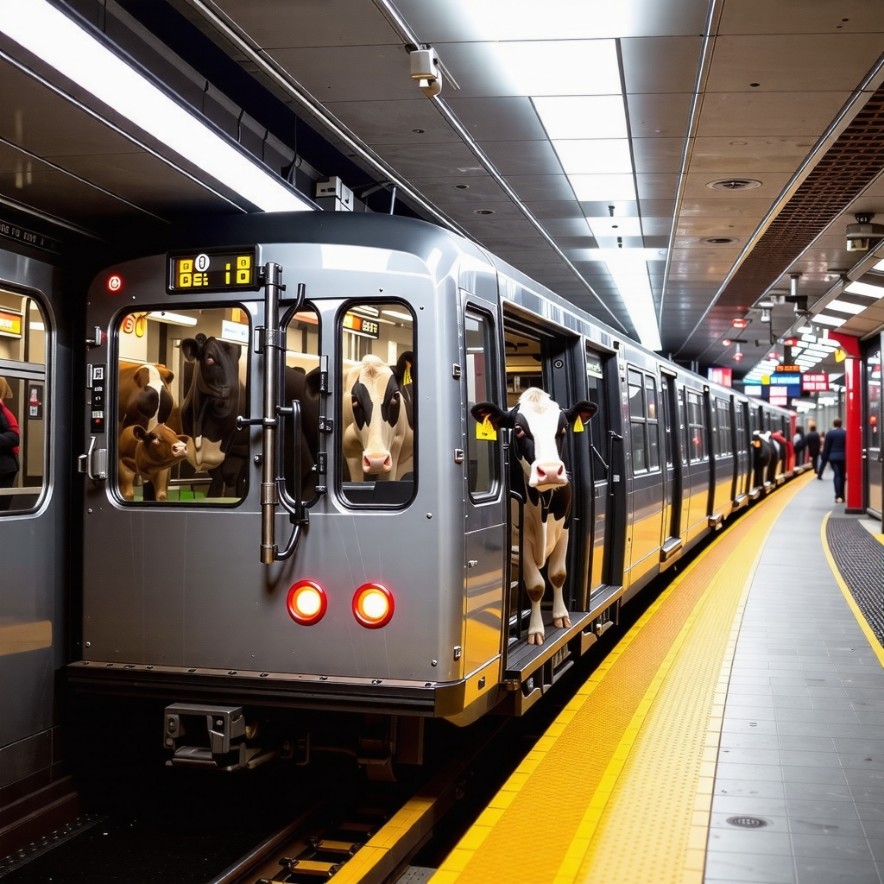}\\
SD3 & SD3 Turbo & SD3 & SD3 Turbo &SD3 & SD3 Turbo\vspace{0.3em}
\end{tabular}
}
\caption{\textbf{Failure cases.} 
While SD3-Turbo retains the image quality of its teacher, prompt alignment can suffer.
Notably, we observe issues such as the merging of distinct entities, diminished accuracy in detailed spatial descriptions, and overlooked negations in prompts, though not universally across different random seeds.
\label{fig:texttoimfailures}
}
\end{figure*}
}

\newcommand{\beps}{\mathbf{\varepsilon}}
\newcommand{\bzero}{\mathbf{0}}

\usepackage{xspace}

\newcommand{\bI}{\mathbf{I}}

\newcommand{\bx}{\mathbf{x}}

\newcommand{\nE}{\mathbb{E}}

\newcommand{\cN}{\mathcal{N}}

\newcommand{\figref}[1]{Fig.~\ref{#1}}
\newcommand{\secref}[1]{Section~\ref{#1}}

\makeatletter
\DeclareRobustCommand\onedot{\futurelet\@let@token\@onedot}
\def\@onedot{\ifx\@let@token.\else.\null\fi\xspace}

\makeatother

\newcommand{\boldparagraph}[1]{\vspace{0.2cm}\noindent{\bf #1} }

\definecolor{darkgreen}{rgb}{0,0.7,0}
\definecolor{darkblue}{RGB}{31,119,180}
\definecolor{darkred}{RGB}{214,39,40}

\usepackage{appendix}

\newlength\savewidth

\renewcommand{\paragraph}[1]{\vspace{1.25mm}\noindent\textbf{#1}}

\usepackage{array}
\newcolumntype{x}[1]{>{\centering\arraybackslash}p{#1pt}}
\newcolumntype{y}[1]{>{\raggedright\arraybackslash}p{#1pt}}
\newcolumntype{z}[1]{>{\raggedleft\arraybackslash}p{#1pt}}

\newcommand{\app}{\raise.17ex\hbox{$\scriptstyle\sim$}}

\definecolor{deemph}{gray}{0.6}

\definecolor{baselinecolor}{gray}{.9}

\usepackage{pifont}%

\usepackage{colortbl}
\usepackage{microtype}
\usepackage{graphicx}
\usepackage{csquotes}
\usepackage{subcaption}
\usepackage{booktabs} %
\usepackage{amsmath}
\usepackage{duckuments}
\usepackage{mathtools}
\usepackage{makecell}
\usepackage{multirow}
\usepackage{comment}
\usepackage{graphbox}
\usepackage{tabularx}
\usepackage{float}
\usepackage{amsfonts}
\usepackage{duckuments}
\usepackage{placeins}
\usepackage[export]{adjustbox} %
\usepackage{pgfplots,pgfplotstable}
\usepgfplotslibrary{fillbetween}
\pgfplotsset{compat=1.16}

\usepackage{fancyvrb}

\RecustomVerbatimCommand{\VerbatimInput}{VerbatimInput}%
{fontsize=\footnotesize,
 frame=lines,  %
 framesep=2em, %
 rulecolor=\color{Gray},
 label=\fbox{\color{Black}Selected Parti Prompts},
 labelposition=topline,
}

\usepackage[capitalize]{cleveref}

\Crefname{section}{Section}{Sections}
\Crefname{section}{Sec.}{Secs.}
\Crefname{table}{Table}{Tables}
\Crefname{table}{Tab.}{Tabs.}
\Crefname{figure}{Figure}{Figures}
\Crefname{figure}{Fig.}{Figures}

\begin{document}

\maketitle
\teaser
\begin{abstract}
Diffusion models are the main driver of progress in image and video synthesis, but suffer from slow inference speed. Distillation methods, like the recently introduced adversarial diffusion distillation (ADD) aim to shift the model from many-shot to single-step inference, albeit at the cost of expensive and difficult optimization due to its reliance on a fixed pretrained DINOv2 discriminator. 
We introduce Latent Adversarial Diffusion Distillation (LADD), a novel distillation approach overcoming the limitations of ADD. In contrast to pixel-based ADD, LADD utilizes generative features from pretrained latent diffusion models. This approach simplifies training and enhances performance, enabling high-resolution  multi-aspect ratio image synthesis. 
We apply LADD to Stable Diffusion 3 (8B) to obtain \emph{SD3-Turbo}, a fast model that matches the performance of state-of-the-art text-to-image generators using only four unguided sampling steps.
Moreover, we systematically investigate its scaling behavior 
and demonstrate LADD's effectiveness in various applications such as image editing and inpainting.
\end{abstract}

\section{Introduction}
\label{sec:intro}

While diffusion models~\cite{SohlDickstein2015DeepUL,ho2020denoising,Song2020ScoreBasedGM,lipman2023flow,liu2022flow}
have revolutionized both synthesis and editing of images~\cite{dhariwal2021diffusion,rombach2022high,ramesh2022hierarchical,balaji2022ediff,saharia2022photorealistic,esser2024scaling,dai2023emu,podell2023sdxl} and videos~\cite{blattmann2023align,blattmann2023stable,singer2022makeavideo,bartal2024lumiere,esser2023structure,ho2022imagen}, 
their iterative nature remains a crucial shortcoming:
At inference, a trained diffusion model usually requires dozens of network evaluations to approximate the probability path from noise to data. This makes sampling 
slow, in particular for large models, 
and limits real-time applications.

Naturally, a large body of work focuses on speeding up the sampling of diffusion models --- both via improved samplers~\cite{song2022denoising,dockhorn2022genie,zhang2023fast,shaul2023bespoke} and distilled models that are trained to match the sample quality of their teacher models in fewer steps~\cite{salimans2022progressive,meng2023distillation,luo2023lcm,song2023consistency}. Very recent distillation works aim at reducing the number of model evaluations to a single step, enabling real-time synthesis~\cite{yin2023onestep,luo2023lcm,xu2023ufogen,sauer2023adversarial,lin2024sdxllightning}.
The best results in the one- and few-step regime are currently achieved with methods that leverage adversarial training~\cite{schmidhuber2020generative,xu2023ufogen,sauer2023adversarial,lin2024sdxllightning}, forcing the output distribution towards the real image manifold. Adversarial Diffusion Distillation (ADD)~\cite{sauer2023adversarial} provides the current state-of-the-art method for single-step synthesis: By leveraging a pretrained DINOv2~\cite{oquab2023dinov2} feature extractor as the backbone of the discriminator, ADD manages to distill SDXL~\cite{podell2023sdxl} into a single-step, real-time text-to-image model.

However, while achieving impressive inference speed, ADD comes with a series of shortcomings: 
First, the usage of the fixed and pretrained DINOv2 network  restricts the discriminator's training resolution to $518 \times 518$ pixels. Furthermore, there is no straightforward way to control the feedback level of the discriminator, e.g., for weighting global shape vs. local features differently. 
Finally, for distilling latent diffusion models, ADD needs to decode to RGB space, as the discriminator has not been trained in latent space, which significantly hinders high-resolution training $> 512^2$ pixels.

More generally, and in contrast to large language models~\cite{kaplan2020scaling,hoffmann2022training} and diffusion models~\cite{peebles2023scalable,esser2024scaling}, current adversarial models do not strictly adhere to scaling laws, and stable training methods usually require extensive hyperparameter tuning. 
In fact, previous attempts at scaling GANs
resulted in diminishing returns when scaling the generator~\cite{sauer2023stylegan,kang2023scaling}. Even more surprisingly, smaller discriminator feature networks often offer better performance than their larger counterparts~\cite{sauer2023adversarial,sauer2023stylegan}. %
These non-intuitive properties are a significant shortcoming for GAN practitioners: Models that follow scaling laws offer predictable improvements in performance, allowing for more strategic and cost-effective scaling, and ultimately better model development.

In this work, we present \emph{Latent Adversarial Diffusion Distillation} (LADD), an approach that offers stable, scalable adversarial distillation of pretrained diffusion transformer models~\cite{peebles2023scalable,esser2024scaling} up to the megapixel regime:
Instead of utilizing discriminative features of, e.g., self-supervised feature networks such as DINOv2, we leverage generative features of a pretrained diffusion model. While directly enabling multi-aspect training, this approach also offers a natural way to control the discriminator features: By targeted sampling of the noise levels during training, we can bias the discriminator features towards more global (high noise level) or local (low noise level) behavior. 
Furthermore, \textit{distillation in latent space} allows for leveraging large student and teacher networks and avoids the expensive decoding step to pixel space, enabling high-resolution image synthesis. 
Consequently, LADD results in a significantly simpler training setup than ADD while outperforming all prior single-step approaches.

We apply LADD to the current state-of-the-art text-to-image model Stable Diffusion 3~\cite{esser2024scaling} and obtain \emph{SD3-Turbo}, a multi-aspect megapixel generator that matches its teacher's image quality in only four sampling steps.
In summary, the core contributions of our work are
\begin{itemize}
    \item \emph{SD3-Turbo}, a fast foundation model supporting high-resolution multi-aspect image generation from text prompts, see Fig.~\ref{fig:teaser} and Fig.~\ref{fig:cherriesbabycherries},  
    \item a significantly simplified distillation formulation that outperforms LADD's predecessor ADD~\cite{sauer2023adversarial} and a systematic study of LADD's scaling behavior, 
    \item a demonstration of the versatility of our approach via two exemplary applications: image editing and image inpainting.
\end{itemize}

We will make code and model weights publicly available.

\cherries

\section{Background}
\label{sec:background}
\subsection{Diffusion Models}
Diffusion models learn to iteratively denoise Gaussian noise $\beps \sim \cN(\bzero, \bI)$ into data. The learnable component in diffusion models is a \emph{denoiser} $D$ that predicts the expected image $\nE[\bx_0 \mid \bx_t, t]$ given a noisy image $\bx_t = \alpha_t \bx_0 + \sigma_t \beps$. While in this work we focus on the rectified flow formulation~\citep{liu2022flow} where $\alpha_t = 1 -t$ and $\sigma_t = t$ for $t \in [0, 1]$, and the denoiser is parameterized as $D(\bx_t, t) = \bx_t - t \cdot F_\theta(\bx_t, t)$, where $F_\theta$ is a large neural network, our method is generally applicable to any diffusion model formalism. The denoiser can be trained via \emph{score matching}~\citep{hyvarinen2005estimation,vincent2011connection},
\begin{align}
    \min_\theta \nE_{\bx_0 \sim p(\bx_0), \beps \sim \cN(\bzero, \bI), t \sim p(t)} \left[\lambda(t) \| D(\bx_t, t) - \bx_0\|_2^2\right],
\end{align}
where  $p(\bx_0)$ is the empirical data distribution, $p(t)$ is a (continuous) distribution over $t \in [0, 1]$ and $\lambda$ is a weighting function. After training, we can generate realistic samples by numerically solving a (stochastic) differential equation (backwards from $t{=}1$ to $t{=}0$)~\citep{Song2020ScoreBasedGM,karras2022elucidating}, iteratively evaluating the learned denoiser $D$.
\subsection{Diffusion Distillation}
While the denoiser $D$ learns to predict clean images with sharp high frequency details for sufficiently small $t$, it also learns to approximate the mean of the empirical data distribution for large $t$, resulting in a highly non-linear differential equation. Therefore, one needs to solve the differential equations with sufficiently small step sizes, resulting in many (expensive) evaluations of the network $F_\theta$.

For many applications, such as text-to-image generation, we are, however, only interested in the final (clean image) distribution at $t{=}0$ which can be obtained from a multitude of different differential equations. In particular, many distillation techniques attempt to learn ``simpler'' differential equations that result in the same distribution at $t{=}0$ however with ``straighter'', more linear, trajectories (which allows for larger step sizes and therefore less evaluations of the network $F_\theta$). Progressive Distillation~\citep{salimans2022progressive}, for example, tries to distill two Euler steps into a single Euler step. This technique iteratively halves the number of steps required, however, it suffers from error accumulation as generally five or more rounds of distillation are needed to obtain a fast model. Reflow~\citep{liu2022flow} is another distillation technique where new models are trained iteratively on synthetic data from older models, and therefore also suffers from error accumulation. In contrast, Consistency Distillation~\citep{song2023consistency} distills models in a single stage without iterative application, however, the training process is quite unstable and requires advanced techniques such as distillation schedules~\citep{song2023consistency}, and extensive hyperparameter tuning. Improved techniques for both Consistency Distillation~\citep{song2023improved, luo2023latent, heek2024multistep,zheng2024trajectory} and Progressive Distillation~\citep{meng2023distillation,lin2024sdxllightning,berthelot2023tract} have since been introduced.
The current top-performing distillation methods for text-to-image applications utilize adversarial training. In particular, Adversarial Diffusion Distillation (ADD)~\cite{sauer2023adversarial}, uses a pretrained feature extractors as its discriminator, achieving performance on par with strong diffusion models such as SDXL~\cite{podell2023sdxl} in only four steps.

\section{Method}
\label{sec:method}

\system 

By leveraging a lower-dimensional latent space, latent diffusion models (LDMs)~\cite{rombach2022high} significantly reduce memory requirements for training, facilitating the efficient scaling of to large model size and high resolutions. This advantage is exemplified by the recently introduced MMDiT family~\cite{esser2024scaling} of LDMs where the largest model (8B parameters) achieves state-of-the art text-to-image synthesis performance.
Our goal is to distill such large LDMs efficiently for high-resolution, multi-aspect image synthesis.
\textit{Latent adversarial diffusion distillation} (LADD), simplifies the distillation process by eliminating the necessity of decoding back to the image space, thereby significantly reducing memory demands in comparison to its predecessor, ADD. 

\boldparagraph{Distillation in latent space.}
An overview of LADD and comparison to ADD is shown in \figref{fig:system}.
In ADD, the ADD-student receives noised input images $x_t$ at the timestep $t$ and generates samples $\hat{x}_\theta (x_t, t)$ aiming to optimize for two objectives: an adversarial loss $L_{adv}$, which involves deceiving a discriminator, and a distillation loss $L_{distill}$, which involves matching the denoised output to that of a frozen DM teacher.
LADD introduces two main modifications: the unification of discriminator and teacher model, and the adoption of synthetic data for training.

\boldparagraph{Unifying teacher and discriminator.}
Instead of decoding and applying a discriminator in image space, we operate exclusively on latents.
First, we renoise the generated latents at timestep $\hat{t}$ drawn from a logit-normal distribution, following~\cite{esser2024scaling}. 
We then apply the teacher model to the noised latents, extracting the full token sequence after each attention block.
On each token sequence, we apply independent discriminator heads. 
Additionally, each discriminator is conditioned on the noise level and pooled CLIP embeddings.

ADD leverages the Projected GAN paradigm~\cite{sauer2021projected}, i.e., applying independent discriminators on features obtained from pretrained features network. 
We can distinguish these feature networks depending on the pretraining task which is either \textit{discriminative} (classification, self-supervised objective) or \textit{generative} (diffusion objective).
Utilizing generative features presents several key benefits over discriminative ones:
\begin{itemize}
    \item \textbf{Efficiency and Simplification.} Generative features eliminate the need for decoding to image space, thereby saving memory and simplifying the overall system compared to ADD. Another possible option is training a discriminative feature network in latent space, yet, discriminative pretraining is non-trivial and top-performing approaches require significant engineering~\cite{caron2021emerging,oquab2023dinov2}.
    \item \textbf{Noise-level specific feedback.} Generative features vary with noise level, providing structured feedback at high noise levels and texture-related feedback at low noise levels~\cite{balaji2022ediff,luo2024diffusion}. By adjusting the parameters of the noise sampling distribution, we gain direct control over discriminator behavior, aligning with the standard practice of loss weighting in diffusion model training~\cite{karras2022elucidating, esser2024scaling}
    \item  \textbf{Multi-Aspect Ratio (MAR).} Since the teacher model is trained on MAR data, it inherently generates relevant features for the discriminators in in this setting.
    \item  
    \textbf{Alignment with Human Perception.} Discriminative models exhibit a notable \textit{texture bias}~\cite{geirhos2018imagenet}, prioritizing texture over global shape, unlike humans who tend to rely on global shape. Jaini~et~al.~\cite{jaini2023intriguing} demonstrates that generative models possess a shape bias closely resembling that of humans and achieve near human-level accuracy on out-of-distribution tasks. This suggests that leveraging pretrained generative features for adversarial training could enhance  alignment with human perception.
\end{itemize}

For the discriminator architecture, we mostly follow~\cite{sauer2023stylegan,sauer2023adversarial}. However, instead of utilizing 1D convolution in the  discriminator, we reshape the token sequence back to its original spatial layout, and transition to 2D convolutions.
Switching from 1D to 2D convolutions circumvents a potential issue in the MAR setting, where a 1D discriminator would process token sequences of varying strides for different aspect ratios, potentially compromising its efficacy.

\boldparagraph{Leveraging synthetic data.}
Classifier-free guidance (CFG)~\cite{ho2022classifier} is essential for generating high-quality samples.
However, in one-shot scenarios, CFG simply oversaturates samples rather than improving text-alignment~\cite{sauer2023stylegan}.
This observation suggests that CFG works best in settings with multiple steps, allowing for corrections of oversaturation issues ins most cases. Additional techniques like dynamic thresholding further ameliorate this issue~\cite{saharia2022photorealistic}.

Text-alignment varies significantly across natural datasets. For instance, while COCO~\cite{lin2014microsoft} images reach an average CLIP 
\footnote{We compute CLIP score using the ViT-g-14 model available at \url{https://github.com/mlfoundations/open_clip}}
score
~\cite{radford2021learning} of 0.29, top-performing diffusion models can achieve notably higher CLIP scores, e.g. SD3 attains a CLIP score of 0.35 on COCO prompts. 
CLIP score is an imperfect metric, yet, the large score differential between natural and synthetic data suggests that generated images are better aligned for a given prompt on average.
To mitigate this issue and avoid additional complexity that is introduced by an auxiliary distillation loss as in ADD, we opt for synthetic data generation via the teacher model at a constant CFG value. 
This strategy ensures high and relatively uniform image-text aligned data and can be considered as an alternative approach for distilling the teacher's knowledge.

As LADD eliminates the need for decoding, we can directly generate latents with the teacher model and omit the additional encoding step for real data. For conditioning of the teacher, we sample prompts from the original training dataset of SD3.

\section{Experiments}
\label{sec:experiments}

In this section, we evaluate our approach in the single-step setting, i.e., starting from pure noise inputs.
For evaluation, we compute the CLIP score on all prompts from DrawBench~\cite{saharia2022photorealistic} and PartiPrompts~\cite{yu2022scaling}.
We train for 10k iterations and the default model for the student, teacher, and data generator is an MMDiT with a depth of 24 ($\sim$2B parameters) if not explicitly stated otherwise. Accordingly, the qualitative outputs in this section are generally of lower quality than the ones of our final (larger) model.

\subsection{Teacher noise distribution}
\figref{fig:sigmaschedules} illustrates the effect of different parametrization for the logit-normal distributions $\pi(t; m, s)$ of the teacher.
When biasing the distribution towards low noise values, we observe missing global coherence while textures and local patches look realistic.
Lacking global coherence is a common problem in adversarial training and additional losses such as classifier or CLIP guidance are often introduced to improve image quality~\cite{sauer2022stylegan,sauer2023stylegan}.
While increasing the bias towards higher noise levels improves coherence, excessively high noise levels can detrimentally affect texture and fine details.
We find $\pi(t; m=1, s=1)$ to be solid choice which we will use for the remainder of this work.
\sigmaschedules

\subsection{Synthetic data}
We aim to answer two questions: Does synthetic data lead to improvements in image-text alignment over real data? And, is an additional distillation loss  $L_{distill}$ necessary?
\figref{fig:distillsynthetic} displays the findings. Training with synthetic data significantly outperforms training with real data. While a distillation loss benefits training with real data, it offers no advantage for synthetic data. Thus, training on synthetic data can be effectively conducted using only an adversarial loss.

\distillsynthetic

\subsection{Latent distillation approaches}
Consistency Distillation~\cite{song2023consistency} is another recent and popular approach for  distillation. 
Latent consistency models (LCM)~\cite{luo2023latent,luo2023lcm} leverage consistency distillation for LDMs where training is conducted exclusively in latent space, similarly to LADD.
For a fair comparison, we train the same student model with LCM and LADD.
We observe much higher volatility for LCM than for LADD training, i.e., outcomes vastly differ for small changes in hyperparameters, different random seeds, and training iterations. For LCM, we run a hyperparameter grid search over the \emph{skipping-step}~\citep{luo2023latent}, noise schedule, and full-finetuning (with and without EMA target~\citep{song2023improved}) vs LoRA-training~\citep{luo2023lcm} and select the best checkpoint out of all runs and over the course of training. For LADD, we train only once and select the last checkpoint.
As~\figref{fig:lcmvladd} shows, LADD outperforms LCM by a large margin.

As discussed in~\secref{sec:background}, Consistency Distillation may require heavy hyperparameter tuning. To the best of our knowledge, we are the first work that attempting LCM training on Diffusion Transformers~\citep{peebles2023scalable,esser2024scaling}, and it may be possible that we have not explore the hyperparameter space well enough. We want to highlight that LCM can potentially achieve more impressive results, as shown by SDXL-LCM~\cite{luo2023lcm,luo2023latent} to which we compare in \secref{sec:text2img}.
We hypothesize that larger models may facilitate LCM training, as evidenced by the substantial improvement when transitioning from SD1.5-LCM to SDXL-LCM~\cite{luo2023latent}.
Nonetheless, our experimental findings indicate that LADD can distill both small and large models effectively and without extensive hyperparameter tuning.

\lcmvladd

\subsection{Scaling Behavior}

We consider three dimension for scaling model size: student, teacher, and data generator.
For the following experiments, we keep two dimensions constant at the default setting (depth=24), allowing variation in just one.
We utilize the models of the scaling study evaluated in~\cite{esser2024scaling}.

\figref{fig:scaling} presents the results. 
Student model size significantly impacts performance, surpassing both data quality and teacher model size in influence. 
Consequently, larger student models do not only demonstrate superior performance as diffusion models~\cite{esser2024scaling}, but that performance advantage is effectively transferred to their distilled versions. 
While teacher models and data quality contribute to improvements, their benefits plateau, indicating diminishing returns beyond certain thresholds. This pattern suggests a strategy for optimizing resource allocation, especially under memory constraints, by prioritizing larger student models while allowing for smaller teacher models without substantially compromising performance.

\scaling

\subsection{Direct preference optimization.}
For better human preference alignment, we finetune our models via \emph{Diffusion DPO} (\citep{wallace2023diffusion}), an adaption of the Direct Preference Optimization~(DPO)~\citep{rafailov2023direct} technique to diffusion models. In particular, we introduce learnable Low-Rank Adaptation (LoRA) matrices (of rank 256) for all linear layers into the teacher model and finetune it for 3k iterations with the DPO objective. 
For the subsequent LADD training, we use the DPO-finetuned model for student, teacher, and data generation.
Interestingly, we find that we can further improve our LADD-student model by reapplying the original DPO-LoRA matrices. The resulting model achieves a win rate of 56\% in a human preference study against the initial, non-DPO LADD-student evaluated at a single step. The human preference study follows the procedures outlined in \secref{supp:preference_study}. DPO is even more impactful in the multi-step setting, as shown in the qualitative examples in \figref{fig:dpo}.

\dpo

\section{Comparison to State-of-the-Art}
\label{sec:sota}

Our evaluations begin with the text-to-image synthesis setting.
We then progress to image-to-image tasks, demonstrating the universal applicability of our distillation approach. 
We adopt a training strategy that incorporates both full and partial noise inputs to enable multi-step inference. For multi-step inference we employ a flow consistency sampler. 

We train across four discrete timesteps $t  \in [1, 0.75, 0.5, 0.25]$. For two- and four-step inference, we found the consistency sampler proposed in~\citep{song2023consistency} to work well. For two step inference, we evaluate the model at $t \in [1, 0.5]$.
At higher resolutions ($>512^2$ pixels), an initial warm-up phase is crucial for training stability; thus, we start with lower noise levels (initial probability distribution $p=[0,0,0.5,0.5]$). After 500 iterations, the focus shifts towards full noise ($p=[0.7,0.1,0.1,0.1]$) to refine single-shot performance. Lastly, MAR training follows the binning strategy outlined in~\cite{podell2023sdxl,esser2024scaling}.

\subsection{Text-to-Image Synthesis}
\label{sec:text2img}

For our main comparison to other approaches, we conduct user preference studies, assessing image quality and prompt alignment, see \secref{supp:preference_study} for details.
\figref{fig:humanevalallsingle} presents the results in the single step setting. SD3-Turbo clearly outperforms all baselines in both image quality and prompt alignment. 
Taking four steps instead of one significantly improves results further which we also illustrate in~\figref{fig:qualitativesteps}. 
We also evaluate SD3-Turbo at four steps against various state-of-the-art text-to-image models in \figref{fig:humanevalallmultiple}.
SD3-Turbo reaches the same image quality as its teacher model SD3, but in four instead of 50 steps.
Although there is a slight reduction in prompt alignment relative to SD3, SD3-Turbo still beats strong baselines like Midjourney~v6. 
We provide high-resolution, multi-aspect samples from SD3-Turbo in Fig.~\ref{fig:teaser} and Fig.~\ref{fig:cherriesbabycherries}.

\humanevalallsingle
\humanevalallmultiple
\qualsteps

\subsection{Image-to-Image Synthesis}
\label{sec:img2img}
It is straightforward to apply LADD to tasks other than text-to-image synthesis.
To validate this claim, we apply LADD to instruction-guided image editing and image inpainting.
We first continue training the pretrained text-to-image diffusion model with the diffusion objective and the dataset adjusted for the respective task. We refer to these models as \textit{SD3-edit} (depth=24) and \textit{SD3-inpainting} (depth=18) respectively.
We then apply LADD as described in Sec.~\ref{sec:method} to distill the image-to-image models, resulting in \textit{SD3-edit Turbo} and \textit{SD3-inpainting Turbo}.

\boldparagraph{Image Editing}
For the image editing task we consider instruction-based editing~\cite{brooks2023instructpix2pix}. Following \cite{brooks2023instructpix2pix,sheynin2023emu}, we condition on the input image via channel-wise concatenation and train on paired data with edit instructions. We use the synthetic InstrucPix2Pix dataset, for which we follow~\cite{boesel2024tiny} and upsample the original $512^2$ pixel samples using SDXL \cite{podell2023sdxl}.
Similar to \cite{sheynin2023emu} we use additional data from bidirectional controlnet tasks (canny edges, keypoints, semantic segmentation, depth maps, HED lines) as well as object segmentation. 
During sampling, we guide the edit model with a nested classifier-free guidance formulation~\cite{ho2022classifier,brooks2023instructpix2pix}, which allows us to utilize different strengths $w$ for the image and text conditioning.

Fig.~\ref{fig:edit_comparison} shows the effectiveness of the distilled model especially for style editing tasks and object swaps by integrating the edited object well with the scene. 
We attribute this improved harmonization capability compared to other approaches to the adversarial loss.
In Fig.~\ref{fig:imgtoimgmetrics} (Left), we plot the trade-off between CLIP image similarity and CLIP image editing direction similarity \cite{radford2021learning,brooks2023instructpix2pix}. We observe that our student model  matches the performance of its teacher in a single step. The notable increase in speed comes at the expense of controllability as the student does not allow to control the trade-off between image and text edit guidance strengths.

\editcomparison
\imgtoimgmetrics
\inpaintcomparison

\boldparagraph{Image Inpainting}
For image inpainting, we condition on the masked input image for which we employ different masking strategies, ranging from narrow strokes, round cutouts and rectangular cutouts to outpainting masks. Furthermore, we always condition on the input image during training and inference, only omitting the text conditioning for the unconditional case.
This configuration differs from that used in the editing task, where we employ the nested classifier-free guidance formulation.
For distillation, we use the same LADD hyperparameters as for the editing model. Since we do not employ synthetic data for this task, we use an additional distillation loss to improve text-alignment.
Our baselines are LaMa~\cite{suvorov2022resolution} and SD1.5-inpainting~\footnote{\url{https://huggingface.co/runwayml/stable-diffusion-inpainting}}.
We sample LaMa and SD1.5-inpainting with the corresponding binary mask. SD3-inpainting is sampled for 50 steps with guidance strength 4, SD1.5 is sampled with the proposed default parameters, i.e., 50 steps, guidance scale 7.5.

\figref{fig:inpaintcomparison} and \figref{fig:imgtoimgmetrics} (Right) present qualitative and quantitative evaluations of the baselines and our model.
Again, our distilled model performs on par with its teacher in a single step. LaMa beats all models on LPIPS, yet the high FID and qualitative comparisons show that LaMa lacks behind when large, non-homogeneous areas are masked.

\section{Limitations}
\label{sec:limitations}

In the human preference study detailed in~\secref{sec:text2img}, we demonstrate that while SD3 Turbo maintains the teacher's image quality within just four steps, it does so at the expense of prompt alignment. This trade-off introduces common text-to-image synthesis challenges such as object duplication and merging, fine-grained spatial prompting, and difficulties with negation. 
These issues, while not unique to our model, underscore a fundamental trade-off between model capacity, prompt alignment, and inference speed; exploring and quantifying this trade-off constitutes an exciting future research direction.

In our evaluation of image editing capabilities, we observe a lack of control  due to the absence of adjustable image and text guidance strengths found in comparative methods~\cite{brooks2023instructpix2pix}. 
A potential solution is deliberately adjusting these parameters during the training phase, coupled with model conditioning on these parameters as proposed in~\cite{luo2023latent}.
Lastly, ins some cases the model exhibits rigidity, i.e., it adheres too closely to the input, rendering large changes challenging.

\texttoimfailures

\begin{ack}
We would like to thank Jonas M\"uller for integrating the synthetic data generation pipeline and Vanessa Sauer for her general support. We also thank Emad Mostaque for his outstanding support of open AI research. 
\end{ack}

\bibliographystyle{abbrvnat}
\bibliography{main}

\clearpage
\begin{appendices}
\section{Details on Human Preference Assessment}
\label{supp:preference_study}
Since standard performance metrics for generative models~\cite{salimans2016improved,heusel2017gans} measure practically relevant quality image aspects as aesthetics and scene composition only insufficiently~\cite{kirstain2023pickapic,podell2023sdxl} we use human evaluation to compare our model against the current state-of-the-art in text-to-image synthesis. Both for ours and the competing models, we generate samples based on prompts from the commonly used PartiPrompts benchmark~\cite{yu2022scaling}.
To focus on more complex prompts, we omit the \textit{Basic} category, which predominantly includes single-word prompts. We then selected every fourth prompt from the remaining categories, resulting in a total of 128 prompts, detailed in \Cref{supp:partiprompts}. For each prompt, we generate four samples per model.

For the results presented in \Cref{fig:humanevalallsingle,fig:humanevalallmultiple} we compare our model with each competing one in a 1v1 comparison, where a human evaluator is presented samples from both models for the same text prompt and forced to pick one of those. To prevent biases, evaluators are restricted from participating in more than one of our studies. For the prompt following task, we show the respective prompt above the two images and ask, \enquote{Which image looks more representative of the text shown above and faithfully follows it?} When probing for visual quality we ask \enquote{Given the prompt above, which image is of higher quality and aesthetically more pleasing?}. When comparing two models we choose the number of human evaluators such that each prompt is shown four times on average for each task    

\subsection{List of Prompts used for Human Evaluation}
\label{supp:partiprompts}
\begin{Verbatim}
"A city in 4-dimensional space-time"
"Pneumonoultramicroscopicsilicovolcanoconiosis"
"A black dog sitting on a wooden chair. 
A white cat with black ears is standing up with its paws on the chair."
"a cat patting a crystal ball with the number 7 written on it in black marker"
"a barred owl peeking out from dense tree branches"
"a cat sitting on a stairway railing"
"a cat drinking a pint of beer"
"a bat landing on a baseball bat"
"a black dog sitting between a bush and a pair of green pants standing up with nobody
inside them"
"a close-up of a blue dragonfly on a daffodil"
"A close-up of two beetles wearing karate uniforms and fighting, jumping over a 
waterfall."
"a basketball game between a team of four cats and a team of three dogs"
"a bird standing on a stick"
"a cat jumping in the air"
"a cartoon of a bear birthday party"
"A cartoon tiger face"
"a dog wearing a baseball cap backwards and writing BONEZ on a chalkboard"
"a basketball to the left of two soccer balls on a gravel driveway"
"a photograph of a fiddle next to a basketball on a ping pong table"
"a black baseball hat with a flame decal on it"
"a doorknocker shaped like a lion's head"
"a coffee mug floating in the sky"
"a bench without any cats on it"
"a harp without any strings"
"long shards of a broken mirror reflecting the eyes of a great horned owl"
"view of a clock tower from above"
"a white flag with a red circle next to a solid blue flag"
"a bookshelf with ten books stacked vertically"
"a comic about two cats doing research"
"a black baseball hat"
"A castle made of cardboard."
"a footprint shaped like a peanut"
"a black t-shirt with the peace sign on it"
"a book with the words 'Don't Panic!' written on it"
"A raccoon wearing formal clothes, wearing a tophat and holding a cane. 
The raccoon is holding a garbage bag. Oil painting in the style of abstract cubism."
"a young badger delicately sniffing a yellow rose, richly textured oil painting"
"A tornado made of bees crashing into a skyscraper. painting in the style of Hokusai."
"a drawing of a house on a mountain"
"a dutch baroque painting of a horse in a field of flowers"
"a glass of orange juice to the right of a plate with buttered toast on it"
"a bloody mary cocktail next to a napkin"
"a can of Spam on an elegant plate"
"A bowl of soup that looks like a monster made out of plasticine"
"A castle made of tortilla chips, in a river made of salsa. 
There are tiny burritos walking around the castle"
"a close-up of a bloody mary cocktail"
"a close-up of an old-fashioned cocktail"
"three green peppers"
"A bowl of Beef Pho"
"a painting of the food of china"
"Two cups of coffee, one with latte art of a heart. The other has latte art of stars."
"Two cups of coffee, one with latte art of yin yang symbol. The other has latte art of 
a heart."
"A set of 2x2 emoji icons with happy, angry, surprised and sobbing faces. 
The emoji icons look like dogs. All of the dogs are wearing blue turtlenecks."
"a kids' book cover with an illustration of white dog driving a red pickup truck"
"a drawing of a series of musical notes wrapped around the Earth"
"a triangle with a smiling face"
"a black background with a large yellow circle and a small red square"
"A green heart with shadow"
"three small yellow boxes"
"A green heart"
"A heart made of water"
"a cute illustration of a horned owl with a graduation cap and diploma"
"a flag with a dinosaur on it"
"G I G G L E painted in thick colorful lettering as graffiti on a faded red brick wall 
with a 
splotch of exploding white paint."
"A high resolution photo of a donkey in a clown costume giving a lecture at the front
of a lecture hall. 
The blackboard has mathematical equations on it. There are many students in the 
lecture hall."
"An empty fireplace with a television above it. The TV shows a lion hugging a giraffe."
"a ceiling fan with an ornate light fixture"
"a small kitchen with a white goat in it"
"a ceiling fan with five brown blades"
"two pianos next to each other"
"a kitchen with a large refrigerator"
"A glass of red wine tipped over on a couch, with a stain that writes "OOPS" on the 
couch."
"the words 'KEEP OFF THE GRASS' written on a brick wall"
"a white bird in front of a dinosaur standing by some trees"
"a beach with a cruise ship passing by"
"a chemtrail passing between two clouds"
"a grand piano next to the net of a tennis court"
"a peaceful lakeside landscape with migrating herd of sauropods"
"a marina without any boats in it"
"a view of the Earth from the moon"
"an aerial photo of a sandy island in the ocean"
"a tennis court with three yellow cones on it"
"a basketball hoop with a large blue ball stuck in it"
"a horse in a field of flowers"
"a cloud in the shape of a elephant"
"a painting of a white country home with a wrap-around porch"
"a store front with 'Grassy Meadow' written on it"
"a black dog jumping up to hug a woman wearing a red sweater"
"a grandmother reading a book to her grandson and granddaughter"
"a child eating a birthday cake near some palm trees"
"a cricket team walking on to the pitch"
"a man riding a cat"
"A boy holds and a girl paints a piece of wood."
"A man gives a woman a laptop and a boy a book."
"a three quarters view of a man getting into a car"
"two people facing the viewer"
"a family of four posing at the Grand Canyon"
"a boy going to school"
"a father and a son playing tennis"
"a cartoon of a man standing under a tree"
"a comic about a family on a road trip"
"a baby daikon radish in a tutu walking a dog"
"a plant growing on the side of a brick wall"
"a flower with a cat's face in the middle"
"a smiling banana wearing a bandana"
"several lily pads without frogs"
"two red flowers and three white flowers"
"a flower with large yellow petals"
"A photo of a four-leaf clover made of water."
"A photo of a palm tree made of water."
"A photograph of the inside of a subway train. There are frogs sitting on the seats. 
One of them is reading a newspaper. The window shows the river in the background."
"a blue airplane taxiing on a runway with the sun behind it"
"a car with tires that have yellow rims"
"a bike rack with some bike locks attached to it but no bicycles"
"a friendly car"
"a subway train with no cows in it"
"an overhead view of a pickup truck with boxes in its flatbed"
"Three-quarters front view of a blue 1977 Ford F-150 coming around a curve 
in a mountain road and looking over a green valley on a cloudy day."
"two motorcycles facing each other"
"A green train is coming down the tracks"
"a cardboard spaceship"
"a crayon drawing of a space elevator"
"a hot air balloon with chameleon logo. the sun is shining and puffy white clouds are 
in the background."
"A close-up high-contrast photo of Sydney Opera House sitting next to Eiffel tower, 
under a blue night sky of roiling energy, exploding yellow stars, and radiating swirls 
of blue"
"a painting of the mona lisa on a white wall"
"A blue Porsche 356 parked in front of a yellow brick wall"
"a diplodocus standing in front of the Millennium Wheel"
"a herd of buffalo stampeding at the Kremlin"
"A smiling sloth is wearing a leather jacket, a cowboy hat, a kilt and a bowtie. The 
sloth is holding a quarterstaff and a big book. The sloth is standing on grass a few 
feet in front of a shiny VW van with flowers painted on it. wide-angle lens from below."
\end{Verbatim}

\end{appendices}

\end{document}